%% file: Patch_NetVLAD.tex
\documentclass[final]{cvpr}

\usepackage[T1]{fontenc}
\usepackage{times}
\usepackage{amsmath,amssymb,amsfonts}
\usepackage{algorithm}
\usepackage{graphicx}
\usepackage{textcomp}
\usepackage{xcolor}
\usepackage{multirow}
\usepackage{comment}
\usepackage{xfrac}
\usepackage{bm}
\usepackage{microtype}
\usepackage{textcomp}
\usepackage[keeplastbox]{flushend}

\DeclareMathOperator*{\argmin}{argmin}

\def\cvprPaperID{4969} %
\def\confYear{CVPR 2021}

\begin{document}

\graphicspath{{figs/}}

\title{Patch-NetVLAD: Multi-Scale Fusion of Locally-Global Descriptors\\for Place Recognition}

\author{Stephen Hausler \qquad Sourav Garg \qquad Ming Xu \qquad Michael Milford \qquad Tobias Fischer\\
QUT Centre for Robotics, Queensland University of Technology\\
{\tt\footnotesize \{s.hausler, s.garg, m22.xu, michael.milford, tobias.fischer\}@qut.edu.au}
}

\maketitle

\AddToShipoutPicture*{%
     \AtTextUpperLeft{%
         \put(0,30){
           \begin{minipage}{\textwidth}
              \footnotesize
              Preprint version; final version available at \url{http://ieeexplore.ieee.org}\\
              IEEE Conference on Computer Vision and Pattern Recognition (CVPR) (2021)\\
              Published by: IEEE
           \end{minipage}}%
     }%
}

\input{0-abstract}
\input{1-introduction}
\input{2-relatedworks}
\input{3-methods}
\input{4-experimentalsetup}
\input{7-conclusions}

{\small
\bibliographystyle{ieee_fullname}
\bibliography{Patch_NetVLAD,sg}
}

\end{document}

\graphicspath{{figs/}}

\renewcommand{\figurename}{Supplementary Figure}
\renewcommand{\tablename}{Supplementary Table}

\title{Patch-NetVLAD: Multi-Scale Fusion of Locally-Global Descriptors\\for Place Recognition\\[0.5cm]<Supplementary Material>}

\author{Stephen Hausler \qquad Sourav Garg \qquad Ming Xu \qquad Michael Milford \qquad Tobias Fischer\\
QUT Centre for Robotics, Queensland University of Technology\\
{\tt\footnotesize \{s.hausler, s.garg, m22.xu, michael.milford, tobias.fischer\}@qut.edu.au}
}

\maketitle

\AddToShipoutPicture*{%
     \AtTextUpperLeft{%
         \put(0,30){
           \begin{minipage}{\textwidth}
              \footnotesize
              Preprint version; final version available at \url{http://ieeexplore.ieee.org}\\
              IEEE Conference on Computer Vision and Pattern Recognition (CVPR) (2021)\\
              Published by: IEEE
           \end{minipage}}%
     }%
}

\input{0-additional-experimental-results}

\input{1-expanded-dataset-descriptions}

{\small
\bibliographystyle{ieee_fullname}
\bibliography{Supp_Patch_NetVLAD,sg}
}

%% file: 0-abstract.tex
\begin{abstract}
Visual Place Recognition is a challenging task for robotics and autonomous systems, which must deal with the twin problems of appearance and viewpoint change in an always changing world. This paper introduces Patch-NetVLAD, which provides a novel formulation for combining the advantages of both local and global descriptor methods by deriving patch-level features from NetVLAD residuals. Unlike the fixed spatial neighborhood regime of existing local keypoint features, our method enables aggregation and matching of deep-learned local features defined over the feature-space grid. We further introduce a multi-scale fusion of patch features that have complementary scales (i.e.~patch sizes) via an integral feature space and show that the fused features are highly invariant to both condition (season, structure, and illumination) and viewpoint (translation and rotation) changes. Patch-NetVLAD outperforms both global and local feature descriptor-based methods with comparable compute, achieving state-of-the-art visual place recognition results on a range of challenging real-world datasets, including winning the Facebook Mapillary Visual Place Recognition Challenge at ECCV2020. It is also adaptable to user requirements, with a speed-optimised version operating over an order of magnitude faster than the state-of-the-art. By combining superior performance with improved computational efficiency in a configurable framework, Patch-NetVLAD is well suited to enhance both stand-alone place recognition capabilities and the overall performance of SLAM systems.
\end{abstract}

%% file: 1-introduction.tex
\vspace{-0.2cm}
\section{Introduction}
Visual Place Recognition (VPR) is a key prerequisite for many robotics and autonomous system applications, both as a stand-alone positioning capability when using a prior map and as a key component of full Simultaneous Localization And Mapping (SLAM) systems. The task can prove challenging because of major changes in appearance, illumination and even viewpoint, and is therefore an area of active research in both the computer vision~\cite{AR2018,Doan_2019_ICCV,Hong_2019_ICCV,Liu_2019_ICCV,Torii2018,Uy_2018_CVPR,Warburg_2020_CVPR} and robotics~\cite{chen2017deep,chen2018learning,chen2017only,Garg2019,khaliq2019holistic,Lowry2016} communities.

\begin{figure}[t!]
    \centering
    \includegraphics[width=0.99\linewidth]{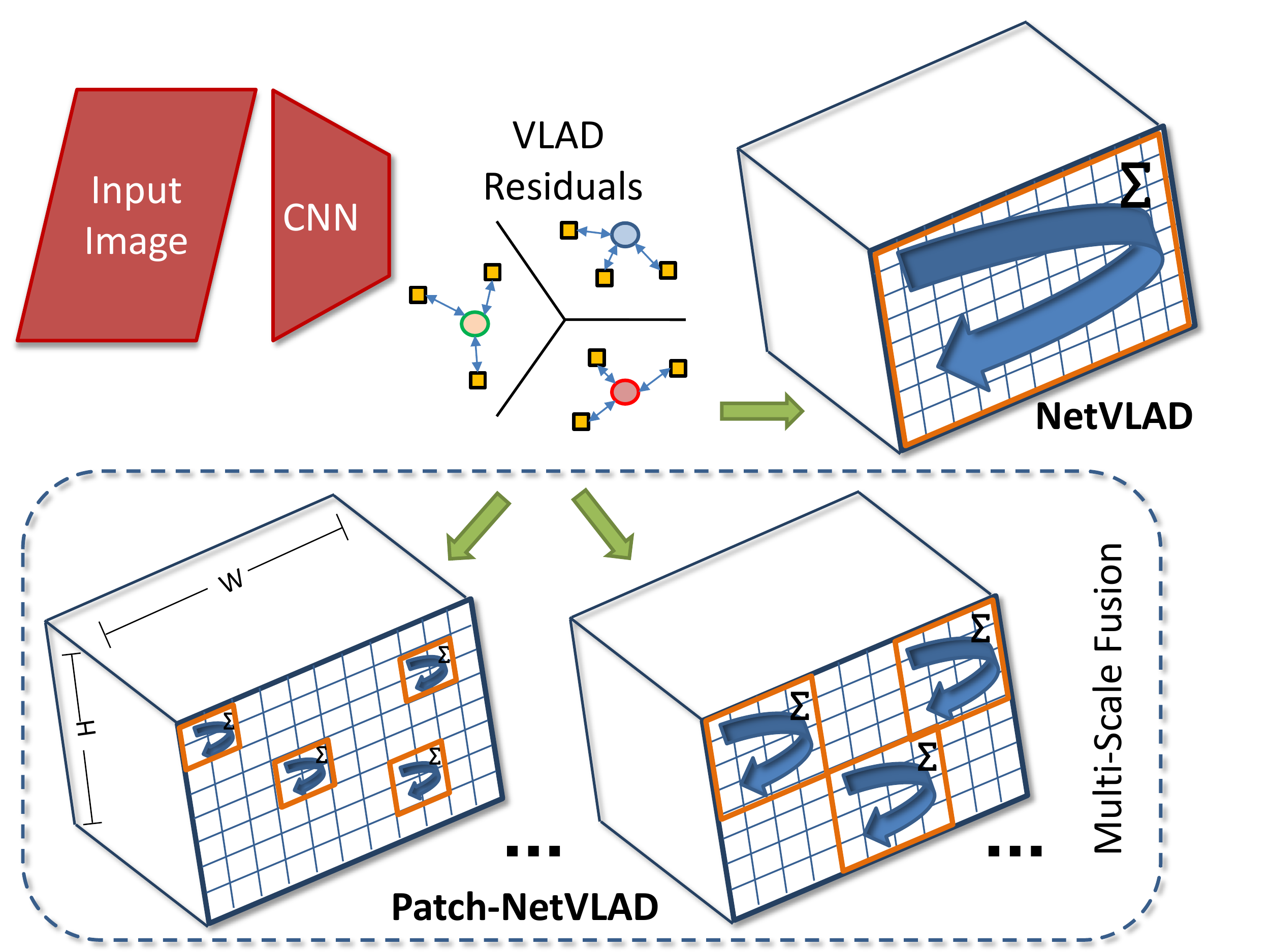}
    \caption{
    \textbf{Patch-NetVLAD} is a novel condition and viewpoint invariant visual place recognition system that produces a similarity score between two images through local matching of \emph{locally-global} descriptors extracted from a set of \emph{patches} in the feature space of each image. Furthermore, by introducing an integral feature space, we are able to derive a multi-scale approach that fuses multiple patch sizes. This is in contrast with the original NetVLAD paper, which performs an appearance only aggregation of the whole feature space into a single global descriptor.}
    \vspace{-0.3cm}
    \label{fig:fancy_figure}
\end{figure}

VPR is typically framed as an image retrieval task~\cite{Lowry2016,revaud2019learning,teichmann2019detect}, where, given a query image, the most similar database image (alongside associated metadata such as the camera pose) is retrieved. There are two common ways to represent the query and reference images: using global descriptors which describe the whole image~\cite{AR2018,Torii2018,Garg2019,revaud2019learning,chen2017deep,radenovic2018fine,Warburg_2020_CVPR}, or using local descriptors that describe areas of interest~\cite{dusmanu2019d2,noh2017large,detone2018superpoint,garg2019look,Cieslewski19threedv}. Global descriptor matching is typically performed using nearest neighbor search between query and reference images. These global descriptors typically excel in terms of their robustness to appearance and illumination changes, as they are directly optimized for place recognition~\cite{AR2018,revaud2019learning}. Conversely, local descriptors are usually cross-matched, followed by geometric verification. Local descriptor techniques prioritize spatial precision, predominantly on a pixel-scale level, using a fixed-size spatial neighborhood to facilitate highly-accurate 6-DoF pose estimation. %
Given the complementary strengths of local and global approaches, there has been little research~\cite{cao2020unifying,schuster2019sdc,teichmann2019detect} attempting to combine them. The novel Patch-NetVLAD system proposed here combines the mutual strengths of local and global approaches while minimizing their weaknesses.

To achieve this goal, we make a number of contributions (see Fig.~\ref{fig:fancy_figure}): First, we introduce a novel place recognition system that generates a similarity score between an image pair through a spatial score obtained through exhaustive matching of \emph{locally-global} descriptors. These descriptors are extracted for densely-sampled local patches within the feature space using a VPR-optimized aggregation technique (in this case NetVLAD~\cite{AR2018}). %
Second, we propose a multi-scale fusion technique that generates and combines these hybrid descriptors of different sizes to achieve improved performance over a single scale approach. 
To minimize the computational growth implications of moving to a multi-scale approach, we develop an integral feature space (analogous to integral images) to derive the local features for varying patch sizes.
Together these contributions provide users with flexibility based on their task requirements: our final contribution is the demonstration of a range of easily implemented system configurations that achieve different performance and computational balances, including a performance-focused configuration that outperforms the state-of-the-art recall performance while being slightly faster, a balanced configuration that performs as well as the state-of-the-art while being $3\times$ faster and a speed-focused configuration that is an order of magnitude faster than the state-of-the-art.

We extensively evaluate the versatility of our proposed system on a large number of well-known datasets~\cite{sunderhauf2013we,RobotCar,Torii2015repetitive,Torii2018,Warburg_2020_CVPR} that capture all challenges present in VPR. We compare Patch-NetVLAD with several state-of-the-art global feature descriptor methods~\cite{AR2018,revaud2019learning,Torii2018}, and additionally introduce new SuperPoint~\cite{detone2018superpoint} and  SuperGlue~\cite{sarlin20superglue}-enabled VPR pipelines as competitive local descriptor baselines. Patch-NetVLAD outperforms the global feature descriptor methods by large margins (from 6\% to 330\% relative increase) across all datasets, and achieves superior performance (up to a relative increase of 54\%) when compared to SuperGlue. %
Patch-NetVLAD won the Facebook Mapillary Long-term Localization Challenge as part of the ECCV 2020 Workshop on Long-Term Visual Localization%
. To characterise the system's properties in detail, we conduct numerous ablation studies showcasing the role of the individual components comprising Patch-NetVLAD, particularly the robustness of the system to changes in various key parameters. 
To foster future research, we make our code available for research purposes: \url{https://github.com/QVPR/Patch-NetVLAD}.

%% file: 2-relatedworks.tex
\section{Related Work}

\textbf{Global Image Descriptors:}
Notable early global image descriptor approaches include aggregation of local keypoint descriptors either through a Bag of Words (BoW) scheme~\cite{sivic2003video, csurka2004visual}, Fisher Vectors (FV)~\cite{jaakkola1999exploiting, perronnin2007fisher} or Vector of Locally Aggregated Descriptors (VLAD)~\cite{jegou2010aggregating, arandjelovic2013all}. Aggregation can be based on either sparse keypoint locations~\cite{sivic2003video,jegou2010aggregating} or dense sampling of an image grid~\cite{Torii2018}. Re-formulating these methods through deep learning-based architectures led to NetVLAD~\cite{AR2018}, NetBoW~\cite{miech2017learnable, ong2018deep} and NetFV~\cite{miech2017learnable}. More recent approaches include ranking-loss based learning~\cite{revaud2019learning}, novel pooling~\cite{radenovic2018fine}, contextual feature reweighting~\cite{FeatReweight2017}, large scale re-training~\cite{Warburg_2020_CVPR}, semantics-guided feature aggregation~\cite{Garg2019,schonberger2018semantic,taira2019right}, use of 3D~\cite{oertel2020augmenting, Uy_2018_CVPR, Liu_2019_ICCV}, additional sensors~\cite{guo2019local, pepperell2014all, Fischer2020} and image appearance translation~\cite{anoosheh2019night,porav2018adversarial}. Place matches obtained through global descriptor matching are often re-ranked using sequential information~\cite{garg2020delta,yin2019mrs,MM2012}, query expansion~\cite{gordo2020attention,chum2011total}, geometric verification~\cite{laskar2020geometric,Garg2019,noh2017large} and feature fusion~\cite{xin2019real, yu2019spatial}. Distinct from existing approaches, this paper introduces Patch-NetVLAD, which reverses the local-to-global process of image description by deriving multi-scale patch features from a global descriptor, NetVLAD.

\textbf{Local Keypoint Descriptors:}
Local keypoint methods are often used to re-rank initial place match candidate lists produced by a global approach~\cite{garg2019look,taira2018inloc,SarlinHFnet}. Traditional hand-crafted local feature methods such as SIFT~\cite{lowe1999object}, SURF~\cite{bay2006surf} and ORB~\cite{rublee2011orb}, and more recent deep-learned local features like LIFT~\cite{yi2016lift}, DeLF~\cite{noh2017large}, SuperPoint~\cite{detone2018superpoint} and D2Net~\cite{dusmanu2019d2}, have been extensively employed for VPR~\cite{noh2017large,cao2020unifying,cummins2008fab}, visual SLAM~\cite{mur2015orb} and 6-DoF localization~\cite{sattler2018benchmarking,toft2020long,dusmanu2019d2,SarlinHFnet}. The two most common approaches of using local features for place recognition are: 1) local aggregation to obtain global image descriptors~\cite{noh2017large} and 2) cross-matching of local descriptors between image pairs~\cite{taira2018inloc}.%

Several learning-based techniques have been proposed for spatially-precise keypoint matching. These include a unified framework for detection, description and orientation estimation~\cite{yi2016lift}; a `describe-then-detect' strategy~\cite{dusmanu2019d2}; multi-layer explicit supervision~\cite{fathy2018hierarchical}; scale-aware negative mining~\cite{spencer2019scale}; and contextual similarity-based unsupervised training~\cite{spencer2020same}. However, the majority of these learning-based methods are optimized for 3D pose estimation, through robust description at a \textit{keypoint} level to improve nearest neighbor matching performance against other keypoint-level descriptors. Local descriptors can be \textit{further} improved by utilizing the larger spatial context, especially beyond the CNN's inherent hierarchical feature pyramid, a key motivation for our approach. 

\textbf{Local Region/Patch Descriptors:}
\cite{sunderhauf2015place} proposed \textit{ConvNet Landmarks} for representing and cross-matching large image regions, explicitly derived from Edge Boxes~\cite{zitnick2014edge}. \cite{chen2017only} discovered landmarks implicitly from CNN activations using mean activation energy of regions defined as `8-connected' feature locations. \cite{camara2020visual} composed region features from CNN activation tensors by concatenating individual spatial elements along the channel dimension. However, these off-the-shelf CNNs or handcrafted region description~\cite{zaffar2020cohog} approaches are not optimized for place recognition, unlike the use of the VPR-trained network in this work.

Learning region descriptors has been studied for specific tasks~\cite{schuster2019sdc,melekhov2019dgc} as well as independently on image patches~\cite{subramaniam2018ncc}. In the context of VPR, \cite{Ge2020self} designed a regions-based self-supervised learning mechanism to improve global descriptors using \textit{image-to-region} matching during training. \cite{zhu2020regional} modeled relations between regions by concatenating local descriptors to learn an improved global image descriptor based on K-Max pooling rather than sum~\cite{babenko2015aggregating} or max pooling (R-MAC)~\cite{tolias2015particular}. \cite{chen2018learning} proposed a `context-flexible' attention mechanism for variable-size regions. However, the learned attention masks were only employed for \textit{viewpoint-assumed} place recognition and could potentially be used for region selection in our proposed VPR pipeline. \cite{teichmann2019detect} proposed \textit{R-VLAD} for describing regions extracted through a trained landmark detector, and combined it with selective match kernels to improve global descriptor matching, thus doing away with cross-region comparisons. \cite{khaliq2019holistic} proposed \textit{RegionVLAD} where region features were defined using average activations of connected components within different layers of CNN feature maps. These region features were then \textit{separately} aggregated as a VLAD representation. Unlike~\cite{khaliq2019holistic,teichmann2019detect}, we remove this separate step by generating region-level VLAD descriptors through NetVLAD, thus reusing the VPR-relevant learned cluster membership of spatial elements.

Existing techniques for multi-scale approaches typically fuse information at the descriptor level, which can lead to loss of complementary or discriminative cues~\cite{xin2019real, yu2019spatial, chen2017deep, neubert2016beyond, gong2014multi, zhu2018attention} due to pooling, or increased descriptor sizes due to concatenation~\cite{le2020city, zhu2020regional,camara2019spatio,camara2020visual}. Distinct from these methods, we consider multi-scale fusion at the final scoring stage, which enables parallel processing with associated speed benefits.

\begin{comment}

\textbf{Local and global descriptors for place recognition and image retrieval:}
NetVLAD~\cite{AR2018}
%
%
%
%
%
%
%

\cite{Warburg_2020_CVPR}: %

DenseVLAD~\cite{Torii2018}

\cite{revaud2019learning}: %

\cite{spencer2020same}
%

\textbf{Place recognition via regions:}

\cite{teichmann2019detect} %

\cite{khaliq2019holistic}%

\cite{camara2020visual}%

\cite{chen2017only}%

\cite{chen2018learning} 
%

%
\cite{Ge2020self}
%

\cite{subramaniam2018ncc}
%

\cite{zhu2020regional}
%

\textbf{Fusing local and global features:}
\cite{cao2020unifying}:%

%

DenseVLAD~\cite{Torii2018} (multi-scale RootSIFT)

\textbf{Multi-scale fusion approaches:}
\cite{chen2017deep}: %

%

%

\cite{Hausler2020}
%

%
%

6-DOF Benchmark TPAMI paper: \cite{toft2020long}

%

%

%

%

%

%

%

%

%

\cite{Hong_2019_ICCV}%

%

\cite{Doan_2019_ICCV}%

\textbf{3D Place Rec}

\cite{Uy_2018_CVPR}:%

\cite{Liu_2019_ICCV}%

\end{comment}

%% file: 3-methods.tex
%
\section{Methodology}
\begin{figure}[t!]
    \centering
    \includegraphics[width=0.99\linewidth]{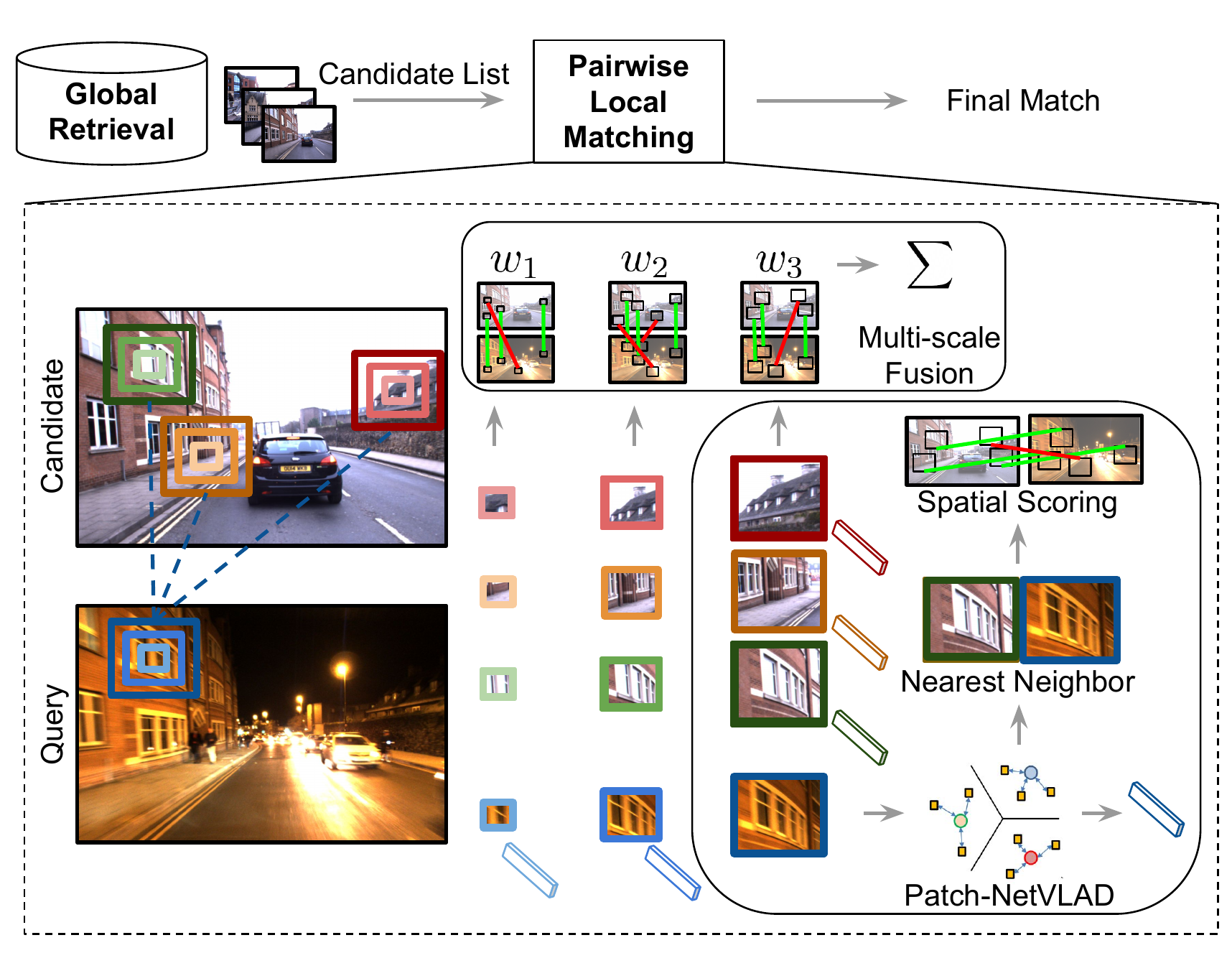}
    \caption{\textbf{Proposed Algorithm Schematic.} Patch-NetVLAD takes as input an initial list of most likely reference matches to a query image, ranked using NetVLAD descriptor comparisons. For top-ranked candidate images, we compute new \textit{locally-global} patch-level descriptors at multiple scales, perform local cross-matching of these descriptors across query and candidate images with geometric verification, and use these match scores to re-order the initial list, producing the final image retrievals.}
    \label{fig:system}
\end{figure}

\emph{Patch-NetVLAD} ultimately produces a similarity score between a pair of images, measuring the spatial and appearance consistency between these images. Our hierarchical approach first uses the original NetVLAD descriptors to retrieve the top-$k$ (we use $k=100$ in our experiments) most likely matches given a query image. We then compute a new type of patch descriptor using an alternative to the VLAD layer used in NetVLAD~\cite{AR2018}, and perform local matching of \textit{patch-level descriptors} to \textit{reorder} the initial match list and refine the final image retrievals. This combined approach minimizes the additional overall computation cost incurred by cross matching patch features without sacrificing recall performance at the final image retrieval stage. An overview of the complete pipeline can be found in Fig.~\ref{fig:system}.

\subsection{Original NetVLAD Architecture}

The original NetVLAD \cite{AR2018} network architecture uses the Vector-of-Locally-Aggregated-Descriptors (VLAD) approach to generate a condition and viewpoint invariant embedding of an image by aggregating the intermediate feature maps extracted from a pre-trained Convolutional Neural Network (CNN) used for image classification~\cite{VGG}. Specifically, let $f_\theta: I \rightarrow \mathbb{R}^{H\times W \times D}$ be the base architecture which given an image $I$, outputs a $H\times W\times D$ dimensional feature map $F$ (\eg the conv5 layer for VGG). The original NetVLAD architecture aggregates these $D$-dimensional features into a $K\times D$-dimensional matrix by summing the residuals between each feature $\mathbf{x}_i\in\mathbb{R}^D$ and $K$ learned cluster centers weighted by soft-assignment. Formally, for $N\times D$-dimensional features, let the VLAD aggregation layer $f_\text{VLAD}:\mathbb{R}^{N \times D} \rightarrow \mathbb{R}^{K \times D}$ be given by
\begin{equation}
    f_\text{VLAD}(F)(j,k)=\sum^N_{i=1} \bar{a}_k(\mathbf{x}_i)(x_i(j) - c_k(j))
\end{equation}
where $x_i(j)$ is the $j^\text{th}$ element of the $i^\text{th}$ descriptor, $\bar{a}_k$ is the soft-assignment function and $\mathbf{c}_k$ denotes the $k^\text{th}$ cluster center. After VLAD aggregation, the resultant matrix is then projected down into a dimensionality reduced vector using a \textit{projection layer} $f_\text{proj}\mkern-8mu:\mathbb{R}^{K\times D} \rightarrow \mathbb{R}^{D_\text{proj}}$ by first applying intra(column)-wise normalization, unrolling into a single vector, L2-normalizing in its entirety and finally applying PCA (learned on a training set) with whitening and L2-normalization. Refer to \cite{AR2018} for more details.

We use this feature-map aggregation method to extract descriptors for \textit{local patches} within the whole feature map $(N \ll H\times W)$ and perform cross-matching of these patches at multiple scales between a query/reference image pair to generate the final similarity score used for image retrieval. This is in contrast to the original NetVLAD paper, which sets $N=H\times W$ and aggregates \textit{all} of the descriptors within the feature map to generate a global image descriptor.

\subsection{Patch-level Global Features}

A core component of our system revolves around extracting global descriptors for densely sampled sub-regions (in the form of \textit{patches}) within the full feature map. We extract a set of $d_{x} \times d_{y}$ patches $\{P_i, x_i, y_i \}_{i=1}^{n_p}$ with stride $s_p$ from the feature map $F\in\mathbb{R}^{H\times W \times D}$, where the total number of patches is given by
\begin{equation}
    n_p=\left \lfloor{\frac{H-d_{y}}{s_p}+1}\right \rfloor  * \left \lfloor{\frac{W-d_{x}}{s_p}+1}\right \rfloor, d_{y},d_{x}\leq H,W
\end{equation}

and $P_i \in \mathbb{R}^{(d_x \times d_y) \times D}$ and  $x_i, y_i$ are the set of patch features and the coordinate of the center of the patch within the feature map, respectively. While our experiments suggest that square patches yielded the best generalized performance across a wide range of environments, future work could consider different patch shapes, especially in specific circumstances (\eg environments with different texture frequencies in the vertical and horizontal directions). %

For each patch, we subsequently extract a descriptor yielding the patch descriptor set $\{\mathbf{f}_i\}_{i=1}^{n_p}$ where $\mathbf{f}_i = f_\text{proj}\left(f_\text{VLAD}\left(P_i\right)\right)\in\mathbb{R}^{D_\text{proj}}$ uses the NetVLAD aggregation and projection layer on the relevant set of patch features.
In all experiments we show how varying the degree of dimensionality reduction on the patch features using PCA can be used to achieve a user-preferred balance of computation time and image retrieval performance (see Section~\ref{subsec:systemconfigs}). We can further improve place recognition performance by extracting patches at multiple scales and observe that using a combination of patch sizes which represent larger sub-regions within the original image improves retrieval (see Section~\ref{subsec:multiregion}). This multi-scale fusion is made computationally efficient using our IntegralVLAD formulation introduced in Section~\ref{ssec:integralvlad}.

Compared to local feature-based matching where features are extracted for comparatively small regions within the image, our patch features implicitly contain \textit{semantic} information about the scene (\eg, building, window, tree) by covering a larger area. We now introduce the remaining parts of our pipeline, which is comprised of mutual nearest neighbor matching of patch descriptors followed by spatial scoring. %

\subsection{Mutual Nearest Neighbours}

Given a set of reference and query features $\{\mathbf{f}_i^r\}_{i=1}^{n_p}$ and $\{\mathbf{f}_i^q\}_{i=1}^{n_p}$, (we assume both images have the same resolution for simplicity), we obtain descriptor pairs from mutual nearest neighbor matches through exhaustive comparison between the two descriptor sets. Formally, let the set of mutual nearest neighbor matches be given by $\mathcal{P}$, where
\begin{equation}
    \mathcal{P} = \left\{\left(i, j\right)\mkern-5mu: i = \text{NN}_r(\mathbf{f}_j^q),\; j = \text{NN}_q(\mathbf{f}_i^r) \right\}
\end{equation}
and $\text{NN}_q(\mathbf{f}) = \argmin_j \|\mathbf{f} - \mathbf{f}_j^q\|_2$ and $\text{NN}_r(\mathbf{f}) = \argmin_j \|\mathbf{f} - \mathbf{f}_j^r\|_2$ retrieve the nearest neighbor descriptor match with respect to Euclidean distance within the query and reference image set, respectively. Given a set of matching patches, we can now compute the spatial matching score used for image retrieval.

\subsection{Spatial Scoring}
\label{subsec:spatialscoring}
We now introduce our spatial scoring methods which yield an image similarity score between a query/reference image pair used for image retrieval. We present two alternatives, a RANSAC-based scoring method which requires more computation time for higher retrieval performance and a spatial scoring method which is substantially faster to compute at the slight expense of image retrieval performance.

\textbf{RANSAC Scoring:}
Our spatial consistency score is given by the number of inliers returned when fitting a homography between the two images, using corresponding patches computed using our mutual nearest neighbor matching step for patch features. We assume each patch corresponds to a 2D image point with coordinates in the center of the patch when fitting the homography. 
We set the error tolerance for the definition of an inlier to be the stride $s_p$. We also normalize our consistency score by the number of patches, which is relevant when combining the spatial score at multiple scales as discussed in Section \ref{subsec:multiregion}.

\textbf{Rapid Spatial Scoring:}
We also propose an alternative to the RANSAC scoring approach which we call \textit{rapid spatial scoring}. This rapid spatial scoring significantly reduces computation time as we can compute this score directly on the matched feature pairs without requiring sampling.%

To compute the rapid spatial score, let $x_d = \{x_i^r - x^q_j\}_{(i, j)\in \mathcal{P}}$ be the set of displacements in the horizontal direction between patch locations for the matched patches, and $y_d$ be the displacements in the vertical direction. In addition, let $\bar{x}_d = \frac{1}{\lvert x_d\rvert} \sum_{x_{d, i}\in x_d} x_{d,i}$ and similarly $\bar{y}_d$ be the mean displacements between matched patch locations. We can then define our spatial score (higher is better) to be
\begin{align}
    s_\text{spatial} = \frac{1}{n_p}\sum_{i\in\mathcal{P}} & \Big(|\max_{j\in\mathcal{P}} x_{d,j}| -  |x_{d,i} - \bar{x}_d| )\Big)^2\nonumber\\
    + & \Big(|\max_{j\in\mathcal{P}} y_{d,j}| - |y_{d,i} - \bar{y}_d| \Big)^2,
\end{align}
where the score comprises the sum of residual displacements from the mean, with respect to the maximum possible spatial offset. The spatial score penalizes large spatial offsets in matched patch locations from the mean offset, in effect measuring the coherency in the overall movement of elements in a scene under viewpoint change.

\subsection{Multiple Patch Sizes}
\label{subsec:multiregion}
We can easily extend our scoring formulation to ensemble patches at multiple scales and further improve performance. For $n_s$ different patch sizes, we can take a convex combination of the spatial matching scores for each patch size as our final matching score. Specifically,

\begin{equation}
    s_\text{spatial} = \sum_{i=1}^{n_s} w_i s_{i, \text{spatial}},
\end{equation}
where $s_{i, \text{spatial}}$ is the spatial score for the $i^{\text{th}}$ patch size and $\sum_i w_i = 1$, $w_i \geq 0$ for all $i$.

\subsection{IntegralVLAD}\label{ssec:integralvlad}

To assist the computation of extracting patch descriptors at multiple scales, we propose a novel \textit{IntegralVLAD} formulation analogous to integral images \cite{crow1984integral}. To see this, note that the aggregated VLAD descriptor (before the projection layer) for a patch can be computed as the sum of all $1\times 1$ patch descriptors each corresponding to a single feature within the patch. This allows us to pre-compute an integral \textit{patch feature map} which can then be used to compute patch descriptors for multi-scale fusion. 
Let the integral feature map $\mathcal{I}$ be given by
\begin{equation}
    \mathcal{I}(i,j) = \sum_{i'<i,j'<j} \mathbf{f}_{i',j'}^1,
\end{equation}
where $\mathbf{f}_{i',j'}^1$ represents the VLAD aggregated patch descriptor (before projection) for a patch size of $1$ at spatial index $i',j'$ in the feature space. We can now recover the patch features for arbitrary scales using the usual approach involving arithmetic over four references within the integral feature map. This is implemented in practice through 2D depth-wise dilated convolutions with kernel $K$, where
\begin{equation}
K =
  \begin{bmatrix}
    1 & -1  \\
    -1 & 1
  \end{bmatrix}
\end{equation}
and the dilation is equal to the required patch size. 

%% file: 4-experimentalsetup.tex
\section{Experimental Results}
\subsection{Implementation}
We implemented Patch-NetVLAD in PyTorch and resize all images to 640 by 480 pixels before extracting our patch features. We train the underlying vanilla NetVLAD feature extractor~\cite{AR2018} on two datasets: Pittsburgh 30k~\cite{Torii2015repetitive} for urban imagery (Pittsburgh and Tokyo datasets), and Mapillary Street Level Sequences~\cite{Warburg_2020_CVPR} for all other conditions. All hyperparameters for training are the same as in~\cite{AR2018}, except for the Mapillary trained model for which we reduced the number of clusters from 64 to 16 for faster training due to the large dataset size.

To find the patch sizes and associated weights, we perform a grid search to find the model configuration that performs best on the RobotCar Seasons v2 training set. This resulted in patch size $d_{x}=d_{y}=5$ (which equates to an 228 by 228 pixel area in the original image) with stride $s{p}=1$ when a single patch size is used, and square patch sizes 2, 5 and 8 with associated weights $w_i = 0.45, 0.15, 0.4$ for the multi-scale fusion. We emphasize that this single configuration is used for \textit{all experiments across all datasets}. While the results will indicate that the proposed system configuration would on average outperform all other methods in a wide range of deployment domains, it is likely that if a highly specialized system was required, further performance increases could be achieved by fine-tuning patch sizes and associated weights for the specific environment.

\subsection{Datasets}
To evaluate Patch-NetVLAD, we used six of the key benchmark datasets: Nordland~\cite{sunderhauf2013we}, Pittsburgh~\cite{Torii2015repetitive}, Tokyo24/7~\cite{Torii2018}, Mapillary Streets~\cite{Warburg_2020_CVPR}, RobotCar Seasons v2~\cite{RobotCar,toft2020long} and Extended CMU Seasons~\cite{CMUSeasons,toft2020long}. Full technical details of their usage are provided in the Supplementary Material; here we provide an overview to facilitate an informed appraisal of the results. Datasets were used in their recommended configuration for benchmarking, including standardized curation (\eg removal of pitch black tunnels and times when the train is stopped for the Nordland dataset~\cite{camara2020visual,SN2015,HauslerHMPF,hausler2019multi}) and use of public validation and withheld test sets where provided (\eg Mapillary).

Collectively the datasets encompass a challenging range of viewpoint and appearance change conditions, partly as a result of significant variations in the acquisition method, including train, car, smartphone and general crowdsourcing. Specific appearance changes are caused by different times of day: dawn, dusk, night; by varying weather: sun, overcast, rain, and snow; and by seasonal change: from summer to winter. Nordland, RobotCar, Extended CMU Seasons and Tokyo 24/7 contain varying degrees of appearance change up to very severe day-night and seasonal changes. The Pittsburgh dataset contains both appearance and viewpoint change, while the MSLS dataset~\cite{Warburg_2020_CVPR} in particular includes simultaneous variations in \emph{all} of the following: geographical diversity (30 major cities across the globe), season, time of day, date (over 7 years), viewpoint, and weather. In total, we evaluate our systems on $\approx$300,000 images. %

\input{5-results-table}
\input{6-ablation-table}

\subsection{Evaluation}
All datasets except for RobotCar Seasons v2 and Extended CMU Seasons are evaluated using the Recall$@N$ metric, whereby a query image is correctly localized if at least one of the top $N$ images is within the ground truth tolerance~\cite{AR2018,Torii2018}. The recall is then the percentage of correctly localized query images, and plots are created by varying $N$.

We deem a query to be correctly localized within the standard ground-truth tolerances for all datasets, \ie 10 frames for Nordland~\cite{hausler2019multi,HauslerHMPF}, 25m translational error for Pittsburgh and Tokyo 24/7~\cite{AR2018}, and 25m translational and 40$^{\circ}$ orientation error for Mapillary~\cite{Warburg_2020_CVPR}.

For RobotCar Seasons v2 and Extended CMU Seasons, we use the default error tolerances~\cite{toft2020long}, namely translational errors of .25, .5 and 5.0 meters and corresponding rotational errors of 2, 5 and 10 degrees. Note that our method is a place recognition system and does not perform explicit 6-DOF pose estimation; the pose estimate for a query image is given by inheriting the pose of the best matched reference image.

\subsection{Comparison to State-of-the-art Methods}
We compare against several benchmark localization solutions: AP-GEM~\cite{revaud2019learning}, DenseVLAD~\cite{Torii2018}, NetVLAD~\cite{AR2018}, and SuperGlue~\cite{sarlin20superglue}. In the recently proposed AP-GEM, the \emph{A}verage \emph{P}recision is directly optimized via  \emph{Ge}neralized \emph{M}ean pooling and a listwise loss. The key idea behind DenseVLAD is to densely sample SIFT features across the image at four different scales, and then aggregate the SIFT features using intra-normalized VLAD.%

\begin{figure}[t!]
    \centering
    \includegraphics[width=0.85\columnwidth]{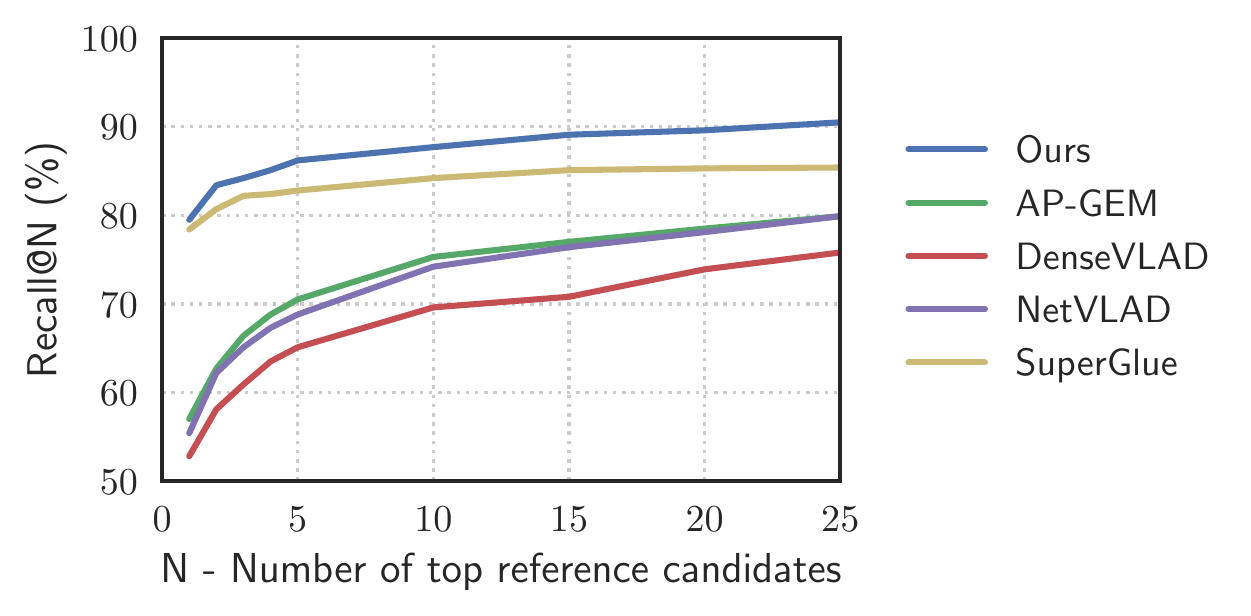}
    \caption{\textbf{Comparison with state-of-the-art.} We show the Recall$@N$ performance of Ours (Multi-RANSAC-Patch-NetVLAD) compared to AP-GEM~\cite{revaud2019learning}, DenseVLAD~\cite{Torii2018}, NetVLAD~\cite{AR2018} and SuperGlue~\cite{sarlin20superglue}, on the Mapillary validation set.}
    \label{fig:recall_at_N_mapillary}
\end{figure}

Finally, we propose an optimistic baseline that utilizes SuperGlue~\cite{sarlin20superglue} as a VPR system. SuperGlue elegantly matches features extracted with SuperPoint~\cite{detone2018superpoint} using a graph neural network that solves an assignment optimization problem. While originally proposed for homography and pose estimation, our proposed SuperGlue VPR baseline achieves what would have been state-of-the-art performance second to Patch-NetVLAD. 
As in Patch-NetVLAD, we provide SuperGlue with the same $k=100$ candidate images extracted using vanilla NetVLAD, and re-rank the candidates by the number of inlier matches. %

Table~\ref{tab:results} and Fig.~\ref{fig:recall_at_N_mapillary} contain quantitative comparisons of Patch-NetVLAD and the baseline methods. Patch-NetVLAD outperforms the best performing global descriptor methods, NetVLAD, DenseVLAD, and AP-GEM, on average by 17.5\%, 14.8\%, and 22.3\% (all percentages stated are absolute differences for R@1) respectively. The differences are particularly pronounced in datasets with large appearance variations, \ie Nordland and Extended CMU Seasons (both seasonal changes), Tokyo 24/7 (including images captured at night time) and RobotCar Seasons as well as Mapillary (both seasonal changes and night time imagery). On the Nordland dataset, the difference between Patch-NetVLAD and the original NetVLAD is 34.5\%.%

A similar trend can be seen when comparing Patch-NetVLAD with SuperGlue. Patch-NetVLAD performs on average 3.1\% better (a relative increase of 6.0\%), which demonstrates that Patch-NetVLAD outperforms a system that benefits from both learned local feature descriptors \emph{and} a learned feature matcher. We hypothesize that Patch-NetVLAD's performance could further benefit from SuperGlue's learned matcher and discuss this opportunity further in Section~\ref{sec:discussion}. SuperGlue is a landmark system and our approach does not beat it in every case, with SuperGlue edging slightly ahead on R@1 and R@5 on Tokyo 24/7 and on R@5 and R@10 on Pittsburgh. Patch-NetVLAD's performance edge is particularly significant when large appearance variations are encountered in \textit{unseen} environments -- not typically used for training local feature methods like SuperGlue (or underlying Superpoint). Thus, Patch-NetVLAD achieves superior performance on Nordland with an \textit{absolute} percentage difference of 15.8\%. Interestingly, the performance difference between Patch-NetVLAD and SuperGlue increases with increasing $N$ -- from 1.1\% for R@1 to 5.1\% for R@25 on the Mapillary dataset (Fig.~\ref{fig:recall_at_N_mapillary}). 

Patch-NetVLAD won the Mapillary Challenge at the ECCV2020 workshops (not yet publicly announced to comply with CVPR's double-blind policy), with Table~\ref{tab:results} showing that Patch-NetVLAD outperformed the baseline method, NetVLAD, by 13.0\% (absolute R@1 increase) on the withheld test dataset. The test set was more challenging than the validation set (48.1\% R@1 and 79.\% R@1 respectively; note that no fine-tuning was performed on any of the datasets), indicating that the Mapillary test set is a good benchmarking target for further research compared to ``near-solved" datasets like Pittsburgh and Tokyo 24/7, where both Patch-NetVLAD and SuperGlue achieve near perfect performance. 

\begin{figure*}[t!]
    \centering
    \includegraphics[width=0.87\textwidth]{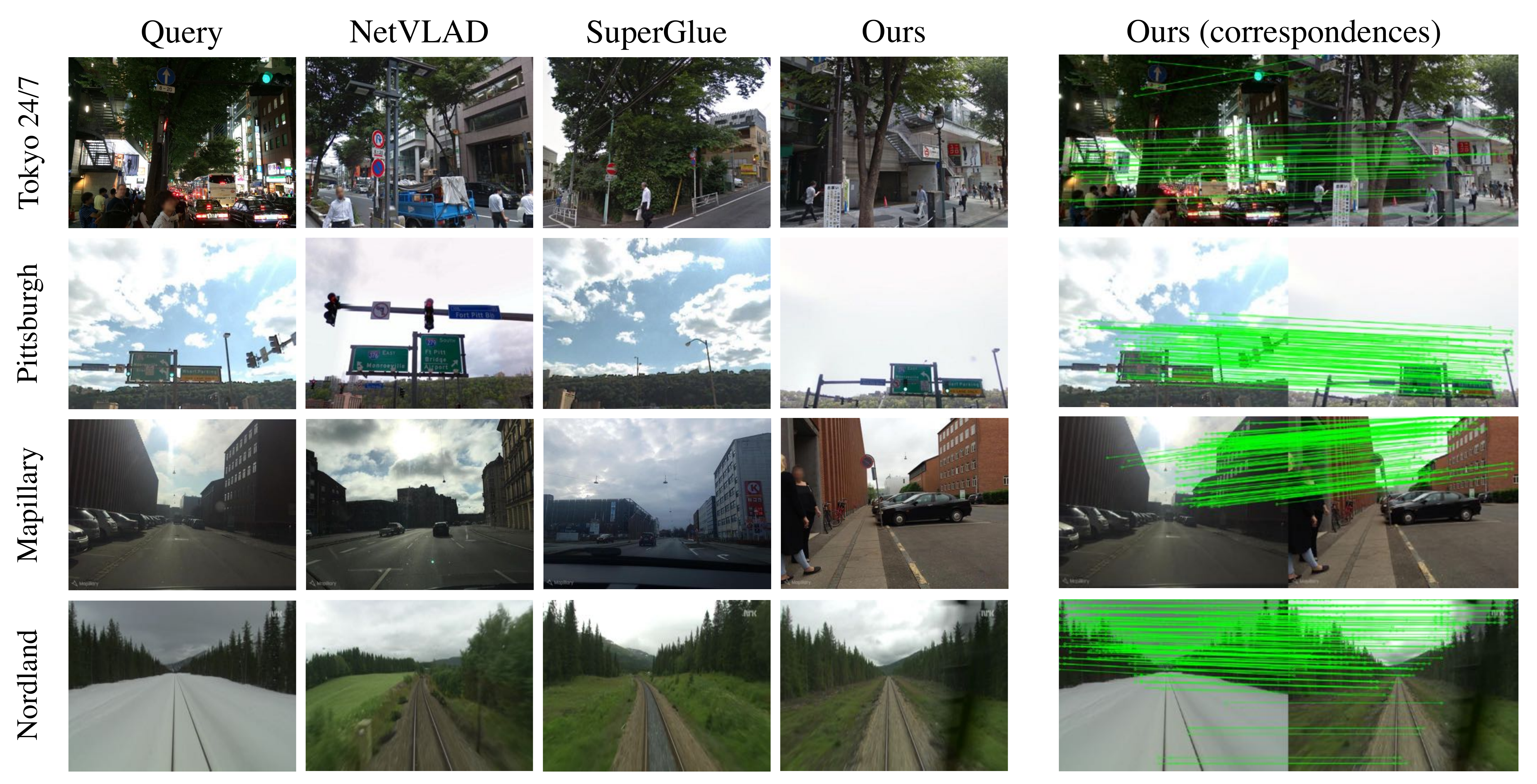}
    \caption{\textbf{Qualitative Results.} In these examples, the proposed Patch-NetVLAD successfully retrieves the matching reference image, while both NetVLAD and SuperGlue produce incorrect place matches. The retrieved image with our approach on the Tokyo 24/7 dataset is a particularly challenging match, with a combination of day vs night-time, severe viewpoint shift and occlusions.}
    \label{fig:qualitative}
\end{figure*}

In Fig.~\ref{fig:qualitative} we show a set of examples images, illustrating the matches retrieved with our method compared to NetVLAD and SuperGlue, along with the patch matches that our algorithm detects.

\subsection{Ablation Studies}
\label{subsec:systemconfigs}

\textbf{Single-scale and Spatial Scoring:}
To analyze the effectiveness of Patch-NetVLAD, we compare with the following variations: 1) Single-RANSAC-Patch-NetVLAD uses a single patch size (\ie 5) instead of multi-scale fusion. 
2) Single-Spatial-Patch-NetVLAD employs a simple but rapid spatial verification method applied to a single patch size (see Section~\ref{subsec:spatialscoring}).
3) Multi-Spatial-Patch-NetVLAD uses the same rapid spatial verification method, however applied to three patch sizes rather than a single patch size as in the previous variant.

The comparison results with these three variations are shown in Table~\ref{tab:results_ablation}. The following numeric results are based on R@1 (recall@1) -- the conclusions generally apply to R@5 and R@10 as well. Our proposed multi-fusion approach (Multi-RANSAC-Patch-NetVLAD) performs on average 2.0\% better than Single-RANSAC-Patch-NetVLAD, demonstrating that a fusion of multiple patch sizes significantly improves task performance. Our approach also provides some compelling options for compute-constrained applications%
; our rapid spatial verification approach is 2.9 times faster on a single patch size (Single-Spatial-Patch-NetVLAD), with only a 0.6\% performance reduction. 
Rapid spatial verification applied to multiple patch sizes (Multi-Spatial-Patch-NetVLAD) is 3.1 times faster, with only a 1.1\% performance degradation. %

\begin{figure}[t!]
    \centering
    \includegraphics[width=0.99\columnwidth]{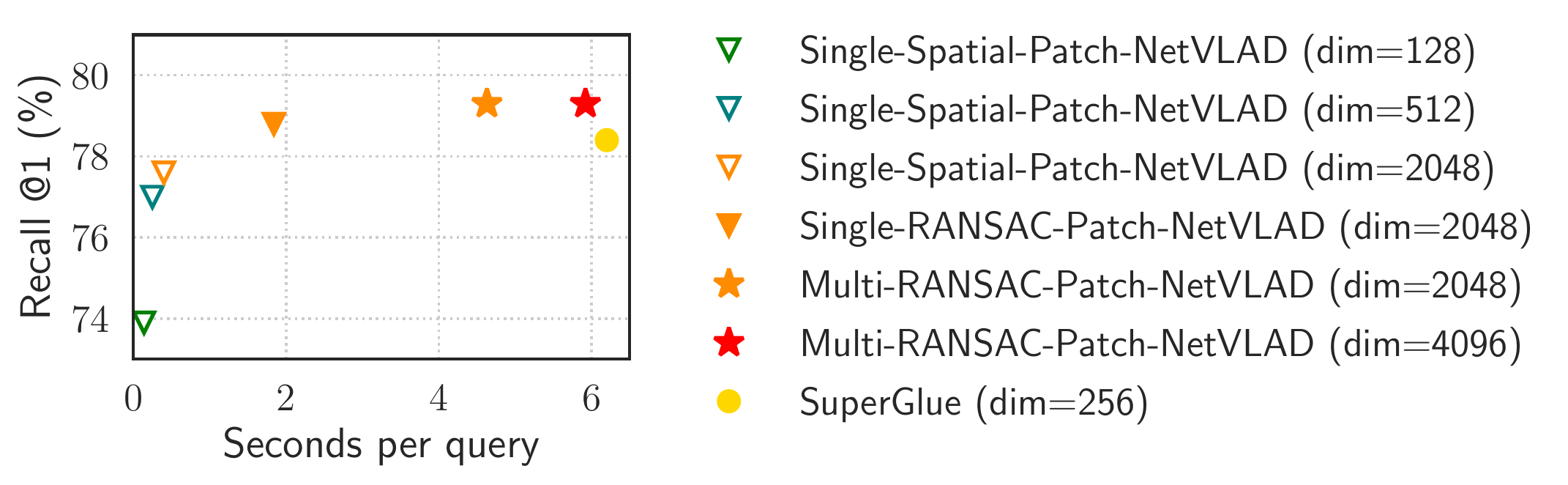}
    \caption{\textbf{Computational time requirements.} The time taken to process one query image is shown on the $x$-axis, with the resulting R@1 shown on the $y$-axis, for the Mapillary dataset. Our pipeline enables a range of system configurations that achieve different performance and computational balances that either outperform or are far faster than current state-of-the-art. %
    }
    \label{fig:compute}
\end{figure}

\textbf{Patch Descriptor Dimension:}
In addition to disabling multi-scale fusion and using our rapid spatial scoring method, the descriptor dimension can be arbitrarily reduced using PCA (as with the original NetVLAD). 
Here, we choose $D_\text{PCA}\mkern-2mu =\mkern-2mu \{128, 512, 2048, 4096\}$. %
Fig.~\ref{fig:compute} shows the number of queries that can be processed per second by various configurations\footnote{These results include both feature extraction and matching times; the Supplementary Material contains further figures that separate feature extraction and feature matching times.}, and the resulting R@1 on the Mapillary validation set. 
Our proposed Multi-RANSAC-Patch-NetVLAD in a performance-focused configuration (red star in Fig.~\ref{fig:compute}) achieves 1.1\% higher recall than SuperGlue (yellow dot) while being slightly (3\%) faster. A balanced configuration (orange triangle) is more than 3 times faster than SuperGlue with comparable performance, while a speed-oriented configuration (blue triangle) is 15 times faster at the expense of just 0.6\% and 1.7\% recall when compared to SuperGlue and our performance-focused configuration respectively. A storage-focused configuration ($D_\text{PCA}\mkern-2mu=\mkern-2mu 128$) still largely outperforms NetVLAD while having similar memory requirements as a SIFT-like descriptor. Our speed-oriented and storage-focused configurations provide practical options for applications like time-critical robotics. Our approach can also run on consumer GPUs, with our performance configuration requiring 7GB GPU memory (batch-size of 5). %

\subsection{Further analysis}
\begin{figure}[t!]
    \centering
    \includegraphics[height=0.36\columnwidth]{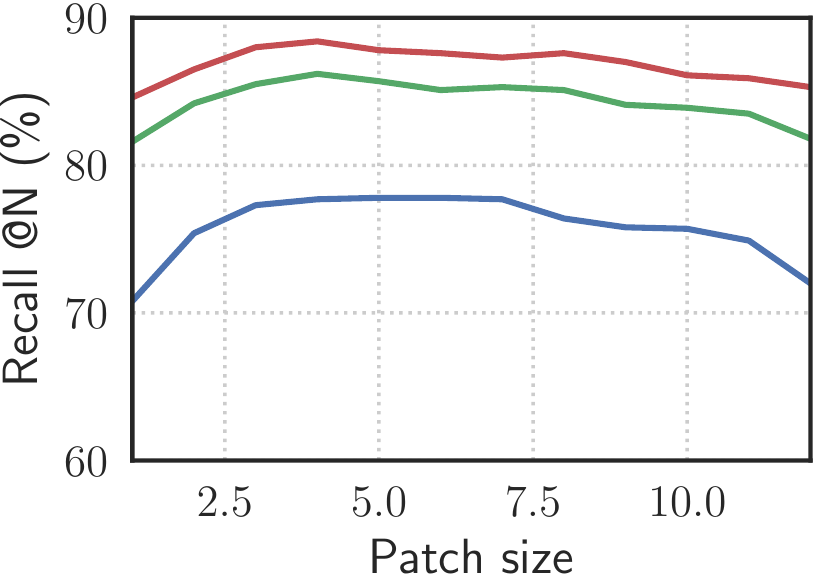}
    \includegraphics[height=0.36\columnwidth]{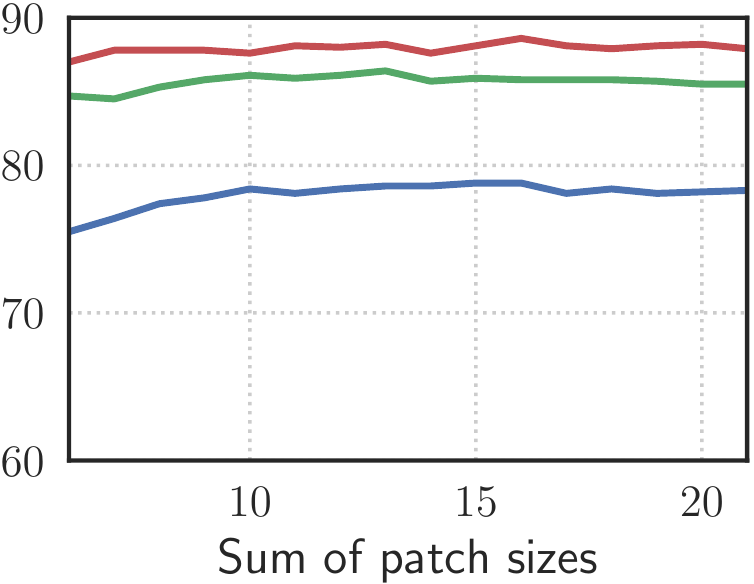}\\
    \includegraphics[width=0.49\columnwidth]{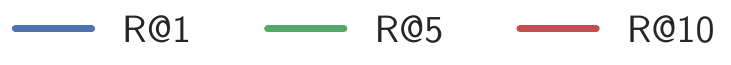}
    \caption{\textbf{Robustness studies for single patch sizes and combined patch sizes.} Left: Recall performance of Single-Patch-NetVLAD with varying patch size, using the Mapillary validation dataset. Performance gradually degrades around a peak at $d_p=4$. The smallest and largest patch sizes perform most poorly, indicating that both local features and global areas are inferior to intermediate size features. An additional issue with large patch sizes is that there are too few patches for effective spatial verification. Right: Recall performance of Multi-Patch-NetVLAD against an indicative measure of cumulative patch dimensions. Our proposed combination of patch dimensions 2, 5 and 8 corresponds to an $x$-axis value of 15; data points to the left show a reduction in the cumulative patch dimension (\eg $\sum_i d_{p,i}=14$ for patch sizes 1, 5 and 8; sizes 2, 4 and 8; and sizes 2, 5 and 7) and so forth; and similarly for increasing patch size combinations to the right. As for variations on the single patch size, performance gracefully degrades around the peak and remains high over a large range.
    }
    \label{fig:robustness}
\end{figure}

We further study the robustness of our approach to the choice of hyperparameters. In Fig.~\ref{fig:robustness} (left) we highlight that Single-Patch-NetVLAD is robust to the choice of the patch size $d_p$: The performance gradually decays from a peak at $d_p=4$. Fig.~\ref{fig:robustness} (right) similarly shows that Patch-NetVLAD is robust to the convex combination of the multi-patch fusion in terms of the patch sizes that are fused. The Supplementary Material provides additional ablation studies, including matching across different patch sizes, complementarity of patch sizes and comparison to other pooling strategies. %

%% file: 5-results-table.tex
\newcount\columncount
\columncount = 22
\input{5-results-table-input-ours}

\input{5-results-table-input-ours-automated}
\input{5-results-table-input-others}

\input{5-results-table-input-rcs-combined}
\input{5-results-table-input-cmu-combined}

\begin{table*}[!t]
  \scriptsize
  \setlength\tabcolsep{0.1cm}
  \renewcommand{\arraystretch}{1.2}
  \centering
  \caption{Quantitative results}%
  \resizebox{0.99\linewidth}{!}{
    \begin{tabular}{c|ccc||ccc||ccc||ccc||ccc||ccc||ccc}
\multirow{2}{*}{Method} & \multicolumn{3}{c||}{\textbf{Nordland}}& \multicolumn{3}{c||}{\textbf{Mapillary (Challenge)}}& \multicolumn{3}{c||}{\textbf{Mapillary (Val. set)}}& \multicolumn{3}{c||}{\textbf{Pittsburgh 30k}}& \multicolumn{3}{c||}{\textbf{Tokyo 24/7}}& \multicolumn{3}{c||}{\textbf{RobotCar Seasons v2}}& \multicolumn{3}{c}{\textbf{Extended CMU Seasons}} \\
\cline{2-\columncount}
& R@1 & R@5 & R@10 & R@1 & R@5 & R@10 & R@1 & R@5 & R@10 & R@1 & R@5 & R@10 & R@1 & R@5 & R@10 & .25m/2\textdegree & .5m/5\textdegree & 5.0m/10\textdegree & .25m/2\textdegree & .5m/5\textdegree & 5.0m/10\textdegree \\
\cline{1-\columncount}
\noalign{\vskip\doublerulesep\vskip-\arrayrulewidth}
\cline{1-\columncount}
AP-GEM~\cite{revaud2019learning} & \ApgemNordlandRone & \ApgemNordlandRfive & \ApgemNordlandRten & \ApgemMapillaryRone & \ApgemMapillaryRfive & \ApgemMapillaryRten & \ApgemMapillaryValRone & \ApgemMapillaryValRfive & \ApgemMapillaryValRten & \ApgemPittsburghRone & \ApgemPittsburghRfive & \ApgemPittsburghRten & \ApgemTokyoTFSRone & \ApgemTokyoTFSRfive & \ApgemTokyoTFSRten & \ApgemRSAllCombinedRone & \ApgemRSAllCombinedRfive & \ApgemRSAllCombinedRten & \ApgemCMUAllCombinedRone & \ApgemCMUAllCombinedRfive & \ApgemCMUAllCombinedRten \\
DenseVLAD~\cite{Torii2018} & \DensevladNordlandRone & \DensevladNordlandRfive & \DensevladNordlandRten & \DensevladMapillaryRone & \DensevladMapillaryRfive & \DensevladMapillaryRten & \DensevladMapillaryValRone & \DensevladMapillaryValRfive & \DensevladMapillaryValRten & \DensevladPittsburghRone & \DensevladPittsburghRfive & \DensevladPittsburghRten & \DensevladTokyoTFSRone & \DensevladTokyoTFSRfive & \DensevladTokyoTFSRten & \DensevladRSAllCombinedRone & \DensevladRSAllCombinedRfive & \DensevladRSAllCombinedRten & \DensevladCMUAllCombinedRone & \DensevladCMUAllCombinedRfive & \DensevladCMUAllCombinedRten \\
NetVLAD~\cite{AR2018} & \NetvladNordlandRone & \NetvladNordlandRfive & \NetvladNordlandRten & \NetvladMapillaryRone & \NetvladMapillaryRfive & \NetvladMapillaryRten & \NetvladMapillaryValRone & \NetvladMapillaryValRfive & \NetvladMapillaryValRten & \NetvladPittsburghRone & \NetvladPittsburghRfive & \NetvladPittsburghRten & \NetvladTokyoTFSRone & \NetvladTokyoTFSRfive & \NetvladTokyoTFSRten & \NetvladRSAllCombinedRone & \NetvladRSAllCombinedRfive & \NetvladRSAllCombinedRten & \NetvladCMUAllCombinedRone & \NetvladCMUAllCombinedRfive & \NetvladCMUAllCombinedRten \\
SuperGlue~\cite{sarlin20superglue} & \SuperglueNordlandRone & \SuperglueNordlandRfive & \SuperglueNordlandRten & \SuperglueMapillaryRone & \SuperglueMapillaryRfive & \SuperglueMapillaryRten & \SuperglueMapillaryValRone & \SuperglueMapillaryValRfive & \SuperglueMapillaryValRten & \SupergluePittsburghRone & \SupergluePittsburghRfive & \SupergluePittsburghRten & \SuperglueTokyoTFSRone & \SuperglueTokyoTFSRfive & \SuperglueTokyoTFSRten & \SuperglueRSAllCombinedRone & \SuperglueRSAllCombinedRfive & \SuperglueRSAllCombinedRten & \SuperglueCMUAllCombinedRone & \SuperglueCMUAllCombinedRfive & \SuperglueCMUAllCombinedRten \\\cline{1-\columncount}
\noalign{\vskip\doublerulesep\vskip-\arrayrulewidth}
\cline{1-\columncount}
\textbf{Ours} & \OursNordlandThreeRone & \OursNordlandThreeRfive & \OursNordlandThreeRten & \OursMapillaryThreeRone & \OursMapillaryThreeRfive & \OursMapillaryThreeRten & \OursMapillaryValThreeRone & \OursMapillaryValThreeRfive & \OursMapillaryValThreeRten & \OursPittsburghThreeRone & \OursPittsburghThreeRfive & \OursPittsburghThreeRten & \OursTokyoTFSThreeRone & \OursTokyoTFSThreeRfive & \OursTokyoTFSThreeRten & \OursThreeRSAllCombinedRone & \OursThreeRSAllCombinedRfive & \OursThreeRSAllCombinedRten & \OursThreeCMUAllCombinedRone & \OursThreeCMUAllCombinedRfive & \OursThreeCMUAllCombinedRten \\
\end{tabular}%
    }
  \label{tab:results}%
  \vspace*{-0.2cm}
\end{table*}%

%% file: 5-results-table-input-ours.tex
\newcommand{\OursMapillaryRone}{\textbf{48.1}}
\newcommand{\OursMapillaryRfive}{\textbf{57.6}}
\newcommand{\OursMapillaryRten}{\textbf{60.5}}

\newcommand{\OursMapillaryThreeRone}{\OursMapillaryRone}
\newcommand{\OursMapillaryThreeRfive}{\OursMapillaryRfive}
\newcommand{\OursMapillaryThreeRten}{\OursMapillaryRten}

\newcommand{\OursAppearanceMapillaryRone}{---}
\newcommand{\OursAppearanceMapillaryRfive}{---}
\newcommand{\OursAppearanceMapillaryRten}{---}

\newcommand{\OursRansacOnlyMapillaryRone}{---}
\newcommand{\OursRansacOnlyMapillaryRfive}{---}
\newcommand{\OursRansacOnlyMapillaryRten}{---}

\newcommand{\OursSpatialApproxMapillaryRone}{---}
\newcommand{\OursSpatialApproxMapillaryRfive}{---}
\newcommand{\OursSpatialApproxMapillaryRten}{---}

\newcommand{\OursMapillaryValRone}{77.8}
\newcommand{\OursMapillaryValRfive}{85.7}
\newcommand{\OursMapillaryValRten}{\textbf{87.8}}

\newcommand{\OursMapillaryValThreeRone}{\textbf{79.5}}
\newcommand{\OursMapillaryValThreeRfive}{\textbf{86.2}}
\newcommand{\OursMapillaryValThreeRten}{\textbf{87.7}}

\newcommand{\OursAppearanceMapillaryValRone}{}
\newcommand{\OursAppearanceMapillaryValRfive}{}
\newcommand{\OursAppearanceMapillaryValRten}{}

\newcommand{\OursRansacOnlyMapillaryValRone}{}
\newcommand{\OursRansacOnlyMapillaryValRfive}{}
\newcommand{\OursRansacOnlyMapillaryValRten}{}

\newcommand{\OursSpatialApproxMapillaryValRone}{}
\newcommand{\OursSpatialApproxMapillaryValRfive}{}
\newcommand{\OursSpatialApproxMapillaryValRten}{}

%% file: 5-results-table-input-ours-automated.tex
\newcommand{\OursRansacOnlyNordlandRone}{42.4}
\newcommand{\OursRansacOnlyNordlandRfive}{48.7}
\newcommand{\OursRansacOnlyNordlandRten}{51.2}
\newcommand{\OursAppearanceNordlandRone}{40.1}
\newcommand{\OursAppearanceNordlandRfive}{47.1}
\newcommand{\OursAppearanceNordlandRten}{49.9}
\newcommand{\OursNordlandRone}{42.4}
\newcommand{\OursNordlandRfive}{48.8}
\newcommand{\OursNordlandRten}{51.2}
\newcommand{\OursSpatialApproxNordlandRone}{42.0}
\newcommand{\OursSpatialApproxNordlandRfive}{48.6}
\newcommand{\OursSpatialApproxNordlandRten}{50.8}

\newcommand{\OursRansacOnlyNordlandThreeRone}{}
\newcommand{\OursRansacOnlyNordlandThreeRfive}{}
\newcommand{\OursRansacOnlyNordlandThreeRten}{}
\newcommand{\OursAppearanceNordlandThreeRone}{}
\newcommand{\OursAppearanceNordlandThreeRfive}{}
\newcommand{\OursAppearanceNordlandThreeRten}{}
\newcommand{\OursNordlandThreeRone}{\textbf{44.9}}
\newcommand{\OursNordlandThreeRfive}{\textbf{50.2}}
\newcommand{\OursNordlandThreeRten}{\textbf{52.2}}
\newcommand{\OursSpatialApproxNordlandThreeRone}{}
\newcommand{\OursSpatialApproxNordlandThreeRfive}{}
\newcommand{\OursSpatialApproxNordlandThreeRten}{}

\newcommand{\OursRansacOnlyPittsburghRone}{87.3}
\newcommand{\OursRansacOnlyPittsburghRfive}{94.2}
\newcommand{\OursRansacOnlyPittsburghRten}{95.7}
\newcommand{\OursAppearancePittsburghRone}{86.9}
\newcommand{\OursAppearancePittsburghRfive}{93.4}
\newcommand{\OursAppearancePittsburghRten}{95.2}
\newcommand{\OursPittsburghRone}{87.3}
\newcommand{\OursPittsburghRfive}{94.2}
\newcommand{\OursPittsburghRten}{95.7}
\newcommand{\OursSpatialApproxPittsburghRone}{87.7}
\newcommand{\OursSpatialApproxPittsburghRfive}{93.8}
\newcommand{\OursSpatialApproxPittsburghRten}{95.4}

\newcommand{\OursRansacOnlyPittsburghThreeRone}{88.3}
\newcommand{\OursRansacOnlyPittsburghThreeRfive}{94.5}
\newcommand{\OursRansacOnlyPittsburghThreeRten}{96.0}
\newcommand{\OursAppearancePittsburghThreeRone}{87.7}
\newcommand{\OursAppearancePittsburghThreeRfive}{93.9}
\newcommand{\OursAppearancePittsburghThreeRten}{95.6}
\newcommand{\OursPittsburghThreeRone}{\textbf{88.7}}
\newcommand{\OursPittsburghThreeRfive}{94.5}
\newcommand{\OursPittsburghThreeRten}{95.9}
\newcommand{\OursSpatialApproxPittsburghThreeRone}{88.4}
\newcommand{\OursSpatialApproxPittsburghThreeRfive}{94.2}
\newcommand{\OursSpatialApproxPittsburghThreeRten}{95.7}

\newcommand{\OursRansacOnlyTokyoTFSRone}{82.2}
\newcommand{\OursRansacOnlyTokyoTFSRfive}{87.3}
\newcommand{\OursRansacOnlyTokyoTFSRten}{89.2}
\newcommand{\OursAppearanceTokyoTFSRone}{73.3}
\newcommand{\OursAppearanceTokyoTFSRfive}{82.2}
\newcommand{\OursAppearanceTokyoTFSRten}{86.0}
\newcommand{\OursTokyoTFSRone}{82.2}
\newcommand{\OursTokyoTFSRfive}{87.3}
\newcommand{\OursTokyoTFSRten}{89.2}
\newcommand{\OursSpatialApproxTokyoTFSRone}{76.8}
\newcommand{\OursSpatialApproxTokyoTFSRfive}{83.8}
\newcommand{\OursSpatialApproxTokyoTFSRten}{86.7}

\newcommand{\OursRansacOnlyTokyoTFSThreeRone}{86.7}
\newcommand{\OursRansacOnlyTokyoTFSThreeRfive}{88.6}
\newcommand{\OursRansacOnlyTokyoTFSThreeRten}{89.5}
\newcommand{\OursAppearanceTokyoTFSThreeRone}{79.4}
\newcommand{\OursAppearanceTokyoTFSThreeRfive}{84.1}
\newcommand{\OursAppearanceTokyoTFSThreeRten}{87.6}
\newcommand{\OursTokyoTFSThreeRone}{86.0}
\newcommand{\OursTokyoTFSThreeRfive}{88.6}
\newcommand{\OursTokyoTFSThreeRten}{\textbf{90.5}}
\newcommand{\OursSpatialApproxTokyoTFSThreeRone}{81.6}
\newcommand{\OursSpatialApproxTokyoTFSThreeRfive}{86.3}
\newcommand{\OursSpatialApproxTokyoTFSThreeRten}{88.3}

%% file: 5-results-table-input-others.tex
\newcommand{\NetvladNordlandRone}{10.4}
\newcommand{\NetvladNordlandRfive}{16.3}
\newcommand{\NetvladNordlandRten}{19.7}

\newcommand{\DensevladNordlandRone}{10.1}
\newcommand{\DensevladNordlandRfive}{17.1}
\newcommand{\DensevladNordlandRten}{21.5}

\newcommand{\ApgemNordlandRone}{5.6}
\newcommand{\ApgemNordlandRfive}{9.1}
\newcommand{\ApgemNordlandRten}{11.2}

\newcommand{\SuperglueNordlandRone}{29.1}
\newcommand{\SuperglueNordlandRfive}{33.4}
\newcommand{\SuperglueNordlandRten}{35.0}

\newcommand{\NetvladMapillaryRone}{35.1}
\newcommand{\NetvladMapillaryRfive}{47.4}
\newcommand{\NetvladMapillaryRten}{51.7}

\newcommand{\DensevladMapillaryRone}{---}
\newcommand{\DensevladMapillaryRfive}{---}
\newcommand{\DensevladMapillaryRten}{---}

\newcommand{\ApgemMapillaryRone}{---}
\newcommand{\ApgemMapillaryRfive}{---}
\newcommand{\ApgemMapillaryRten}{---}

\newcommand{\SuperglueMapillaryRone}{---}
\newcommand{\SuperglueMapillaryRfive}{---}
\newcommand{\SuperglueMapillaryRten}{---}

\newcommand{\NetvladMapillaryValRone}{60.8}
\newcommand{\NetvladMapillaryValRfive}{74.3}
\newcommand{\NetvladMapillaryValRten}{79.5}

\newcommand{\DensevladMapillaryValRone}{52.8}
\newcommand{\DensevladMapillaryValRfive}{65.1}
\newcommand{\DensevladMapillaryValRten}{69.6}

\newcommand{\ApgemMapillaryValRone}{57.0}
\newcommand{\ApgemMapillaryValRfive}{70.5}
\newcommand{\ApgemMapillaryValRten}{75.3}

\newcommand{\SuperglueMapillaryValRone}{78.4}
\newcommand{\SuperglueMapillaryValRfive}{82.8}
\newcommand{\SuperglueMapillaryValRten}{84.2}

\newcommand{\NetvladOxfordDNRone}{33.3}
\newcommand{\NetvladOxfordDNRfive}{51.4}
\newcommand{\NetvladOxfordDNRten}{59.8}

\newcommand{\DensevladOxfordDNRone}{48.8}
\newcommand{\DensevladOxfordDNRfive}{63.7}
\newcommand{\DensevladOxfordDNRten}{70.7}

\newcommand{\ApgemOxfordDNRone}{24.4}
\newcommand{\ApgemOxfordDNRfive}{39.7}
\newcommand{\ApgemOxfordDNRten}{47.0}

\newcommand{\SuperglueOxfordDNRone}{80.0}
\newcommand{\SuperglueOxfordDNRfive}{83.9}
\newcommand{\SuperglueOxfordDNRten}{84.5}

\newcommand{\NetvladPittsburghRone}{83.5}
\newcommand{\NetvladPittsburghRfive}{91.3}
\newcommand{\NetvladPittsburghRten}{94.0}

\newcommand{\DensevladPittsburghRone}{77.7}
\newcommand{\DensevladPittsburghRfive}{88.3}
\newcommand{\DensevladPittsburghRten}{91.6}

\newcommand{\ApgemPittsburghRone}{75.3}
\newcommand{\ApgemPittsburghRfive}{89.3}
\newcommand{\ApgemPittsburghRten}{92.5}

\newcommand{\SupergluePittsburghLargeRone}{91.4}
\newcommand{\SupergluePittsburghLargeRfive}{96.5}
\newcommand{\SupergluePittsburghLargeRten}{97.3}

\newcommand{\SupergluePittsburghRone}{\textbf{88.7}}
\newcommand{\SupergluePittsburghRfive}{\textbf{95.1}}
\newcommand{\SupergluePittsburghRten}{\textbf{96.4}}

\newcommand{\NetvladTokyoTFSRone}{64.8}
\newcommand{\NetvladTokyoTFSRfive}{78.4}
\newcommand{\NetvladTokyoTFSRten}{81.6}

\newcommand{\DensevladTokyoTFSRone}{59.4}
\newcommand{\DensevladTokyoTFSRfive}{67.3}
\newcommand{\DensevladTokyoTFSRten}{72.1}

\newcommand{\ApgemTokyoTFSRone}{40.3}
\newcommand{\ApgemTokyoTFSRfive}{55.6}
\newcommand{\ApgemTokyoTFSRten}{65.4}

\newcommand{\SuperglueTokyoTFSRone}{\textbf{88.2}}
\newcommand{\SuperglueTokyoTFSRfive}{\textbf{90.2}}
\newcommand{\SuperglueTokyoTFSRten}{90.2}

%% file: 5-results-table-input-rcs-combined.tex
\newcommand{\NetvladRSDayCombinedRone}{4.2}
\newcommand{\NetvladRSDayCombinedRfive}{61.3}
\newcommand{\NetvladRSDayCombinedRten}{80.7}

\newcommand{\DensevladRSDayCombinedRone}{5.7}
\newcommand{\DensevladRSDayCombinedRfive}{68.1}
\newcommand{\DensevladRSDayCombinedRten}{85.0}

\newcommand{\ApgemRSDayCombinedRone}{3.3}
\newcommand{\ApgemRSDayCombinedRfive}{50.5}
\newcommand{\ApgemRSDayCombinedRten}{72.5}

\newcommand{\SuperglueRSDayCombinedRone}{5.9}
\newcommand{\SuperglueRSDayCombinedRfive}{71.8}
\newcommand{\SuperglueRSDayCombinedRten}{88.6}

\newcommand{\OursRSDayCombinedRone}{6.3}
\newcommand{\OursRSDayCombinedRfive}{73.2}
\newcommand{\OursRSDayCombinedRten}{88.7}

\newcommand{\OursThreeRSDayCombinedRone}{\textbf{6.7}}
\newcommand{\OursThreeRSDayCombinedRfive}{\textbf{74.0}}
\newcommand{\OursThreeRSDayCombinedRten}{\textbf{89.1}}

\newcommand{\OursAppearanceRSDayCombinedRone}{}
\newcommand{\OursAppearanceRSDayCombinedRfive}{}
\newcommand{\OursAppearanceRSDayCombinedRten}{}

\newcommand{\OursRansacOnlyRSDayCombinedRone}{}
\newcommand{\OursRansacOnlyRSDayCombinedRfive}{}
\newcommand{\OursRansacOnlyRSDayCombinedRten}{}

\newcommand{\OursSpatialApproxRSDayCombinedRone}{}
\newcommand{\OursSpatialApproxRSDayCombinedRfive}{}
\newcommand{\OursSpatialApproxRSDayCombinedRten}{}

\newcommand{\NetvladRSNightCombinedRone}{0.1}
\newcommand{\NetvladRSNightCombinedRfive}{2.4}
\newcommand{\NetvladRSNightCombinedRten}{5.5}

\newcommand{\DensevladRSNightCombinedRone}{0.6}
\newcommand{\DensevladRSNightCombinedRfive}{13.9}
\newcommand{\DensevladRSNightCombinedRten}{23.5}

\newcommand{\ApgemRSNightCombinedRone}{0.1}
\newcommand{\ApgemRSNightCombinedRfive}{1.6}
\newcommand{\ApgemRSNightCombinedRten}{5.4}

\newcommand{\SuperglueRSNightCombinedRone}{0.7}
\newcommand{\SuperglueRSNightCombinedRfive}{\textbf{17.0}}
\newcommand{\SuperglueRSNightCombinedRten}{\textbf{29.9}}

\newcommand{\OursRSNightCombinedRone}{0.4}
\newcommand{\OursRSNightCombinedRfive}{13.5}
\newcommand{\OursRSNightCombinedRten}{24.7}

\newcommand{\OursThreeRSNightCombinedRone}{\textbf{0.8}}
\newcommand{\OursThreeRSNightCombinedRfive}{15.7}
\newcommand{\OursThreeRSNightCombinedRten}{27.9}

\newcommand{\OursAppearanceRSNightCombinedRone}{}
\newcommand{\OursAppearanceRSNightCombinedRfive}{}
\newcommand{\OursAppearanceRSNightCombinedRten}{}

\newcommand{\OursRansacOnlyRSNightCombinedRone}{}
\newcommand{\OursRansacOnlyRSNightCombinedRfive}{}
\newcommand{\OursRansacOnlyRSNightCombinedRten}{}

\newcommand{\OursSpatialApproxRSNightCombinedRone}{}
\newcommand{\OursSpatialApproxRSNightCombinedRfive}{}
\newcommand{\OursSpatialApproxRSNightCombinedRten}{}

\newcommand{\NetvladRSAllCombinedRone}{6.5}
\newcommand{\NetvladRSAllCombinedRfive}{23.8}
\newcommand{\NetvladRSAllCombinedRten}{77.7}

\newcommand{\DensevladRSAllCombinedRone}{7.4}
\newcommand{\DensevladRSAllCombinedRfive}{28.1}
\newcommand{\DensevladRSAllCombinedRten}{79.8}

\newcommand{\ApgemRSAllCombinedRone}{4.5}
\newcommand{\ApgemRSAllCombinedRfive}{16.9}
\newcommand{\ApgemRSAllCombinedRten}{62.7}

\newcommand{\SuperglueRSAllCombinedRone}{8.3}
\newcommand{\SuperglueRSAllCombinedRfive}{32.4}
\newcommand{\SuperglueRSAllCombinedRten}{89.9}

\newcommand{\OursRSAllCombinedRone}{8.7}
\newcommand{\OursRSAllCombinedRfive}{31.6}
\newcommand{\OursRSAllCombinedRten}{88.3}

\newcommand{\OursThreeRSAllCombinedRone}{\textbf{9.6}}
\newcommand{\OursThreeRSAllCombinedRfive}{\textbf{35.3}}
\newcommand{\OursThreeRSAllCombinedRten}{\textbf{90.9}}

\newcommand{\OursAppearanceRSAllCombinedRone}{}
\newcommand{\OursAppearanceRSAllCombinedRfive}{}
\newcommand{\OursAppearanceRSAllCombinedRten}{}

\newcommand{\OursRansacOnlyRSAllCombinedRone}{}
\newcommand{\OursRansacOnlyRSAllCombinedRfive}{}
\newcommand{\OursRansacOnlyRSAllCombinedRten}{}

\newcommand{\OursSpatialApproxRSAllCombinedRone}{}
\newcommand{\OursSpatialApproxRSAllCombinedRfive}{}
\newcommand{\OursSpatialApproxRSAllCombinedRten}{}

%% file: 5-results-table-input-cmu-combined.tex
\newcommand{\NetvladCMUParkCombinedRone}{}
\newcommand{\NetvladCMUParkCombinedRfive}{}
\newcommand{\NetvladCMUParkCombinedRten}{}

\newcommand{\DensevladCMUParkCombinedRone}{}
\newcommand{\DensevladCMUParkCombinedRfive}{}
\newcommand{\DensevladCMUParkCombinedRten}{}

\newcommand{\ApgemCMUParkCombinedRone}{}
\newcommand{\ApgemCMUParkCombinedRfive}{}
\newcommand{\ApgemCMUParkCombinedRten}{}

\newcommand{\SuperglueCMUParkCombinedRone}{}
\newcommand{\SuperglueCMUParkCombinedRfive}{}
\newcommand{\SuperglueCMUParkCombinedRten}{}

\newcommand{\OursCMUParkCombinedRone}{}
\newcommand{\OursCMUParkCombinedRfive}{}
\newcommand{\OursCMUParkCombinedRten}{}

\newcommand{\OursThreeCMUParkCombinedRone}{}
\newcommand{\OursThreeCMUParkCombinedRfive}{}
\newcommand{\OursThreeCMUParkCombinedRten}{}

\newcommand{\OursAppearanceCMUParkCombinedRone}{}
\newcommand{\OursAppearanceCMUParkCombinedRfive}{}
\newcommand{\OursAppearanceCMUParkCombinedRten}{}

\newcommand{\OursRansacOnlyCMUParkCombinedRone}{}
\newcommand{\OursRansacOnlyCMUParkCombinedRfive}{}
\newcommand{\OursRansacOnlyCMUParkCombinedRten}{}

\newcommand{\OursSpatialApproxCMUParkCombinedRone}{}
\newcommand{\OursSpatialApproxCMUParkCombinedRfive}{}
\newcommand{\OursSpatialApproxCMUParkCombinedRten}{}

\newcommand{\NetvladCMUUrbanCombinedRone}{}
\newcommand{\NetvladCMUUrbanCombinedRfive}{}
\newcommand{\NetvladCMUUrbanCombinedRten}{}

\newcommand{\DensevladCMUUrbanCombinedRone}{}
\newcommand{\DensevladCMUUrbanCombinedRfive}{}
\newcommand{\DensevladCMUUrbanCombinedRten}{}

\newcommand{\ApgemCMUUrbanCombinedRone}{}
\newcommand{\ApgemCMUUrbanCombinedRfive}{}
\newcommand{\ApgemCMUUrbanCombinedRten}{}

\newcommand{\SuperglueCMUUrbanCombinedRone}{}
\newcommand{\SuperglueCMUUrbanCombinedRfive}{}
\newcommand{\SuperglueCMUUrbanCombinedRten}{}

\newcommand{\OursCMUUrbanCombinedRone}{}
\newcommand{\OursCMUUrbanCombinedRfive}{}
\newcommand{\OursCMUUrbanCombinedRten}{}

\newcommand{\OursThreeCMUUrbanCombinedRone}{}
\newcommand{\OursThreeCMUUrbanCombinedRfive}{}
\newcommand{\OursThreeCMUUrbanCombinedRten}{}

\newcommand{\OursAppearanceCMUUrbanCombinedRone}{}
\newcommand{\OursAppearanceCMUUrbanCombinedRfive}{}
\newcommand{\OursAppearanceCMUUrbanCombinedRten}{}

\newcommand{\OursRansacOnlyCMUUrbanCombinedRone}{}
\newcommand{\OursRansacOnlyCMUUrbanCombinedRfive}{}
\newcommand{\OursRansacOnlyCMUUrbanCombinedRten}{}

\newcommand{\OursSpatialApproxCMUUrbanCombinedRone}{}
\newcommand{\OursSpatialApproxCMUUrbanCombinedRfive}{}
\newcommand{\OursSpatialApproxCMUUrbanCombinedRten}{}

\newcommand{\NetvladCMUSuburbanCombinedRone}{}
\newcommand{\NetvladCMUSuburbanCombinedRfive}{}
\newcommand{\NetvladCMUSuburbanCombinedRten}{}

\newcommand{\DensevladCMUSuburbanCombinedRone}{}
\newcommand{\DensevladCMUSuburbanCombinedRfive}{}
\newcommand{\DensevladCMUSuburbanCombinedRten}{}

\newcommand{\ApgemCMUSuburbanCombinedRone}{}
\newcommand{\ApgemCMUSuburbanCombinedRfive}{}
\newcommand{\ApgemCMUSuburbanCombinedRten}{}

\newcommand{\SuperglueCMUSuburbanCombinedRone}{}
\newcommand{\SuperglueCMUSuburbanCombinedRfive}{}
\newcommand{\SuperglueCMUSuburbanCombinedRten}{}

\newcommand{\OursCMUSuburbanCombinedRone}{}
\newcommand{\OursCMUSuburbanCombinedRfive}{}
\newcommand{\OursCMUSuburbanCombinedRten}{}

\newcommand{\OursThreeCMUSuburbanCombinedRone}{}
\newcommand{\OursThreeCMUSuburbanCombinedRfive}{}
\newcommand{\OursThreeCMUSuburbanCombinedRten}{}

\newcommand{\OursAppearanceCMUSuburbanCombinedRone}{}
\newcommand{\OursAppearanceCMUSuburbanCombinedRfive}{}
\newcommand{\OursAppearanceCMUSuburbanCombinedRten}{}

\newcommand{\OursRansacOnlyCMUSuburbanCombinedRone}{}
\newcommand{\OursRansacOnlyCMUSuburbanCombinedRfive}{}
\newcommand{\OursRansacOnlyCMUSuburbanCombinedRten}{}

\newcommand{\OursSpatialApproxCMUSuburbanCombinedRone}{}
\newcommand{\OursSpatialApproxCMUSuburbanCombinedRfive}{}
\newcommand{\OursSpatialApproxCMUSuburbanCombinedRten}{}

\newcommand{\NetvladCMUAllCombinedRone}{5.8}
\newcommand{\NetvladCMUAllCombinedRfive}{17.9}
\newcommand{\NetvladCMUAllCombinedRten}{78.3}

\newcommand{\DensevladCMUAllCombinedRone}{8.2}
\newcommand{\DensevladCMUAllCombinedRfive}{25.4}
\newcommand{\DensevladCMUAllCombinedRten}{82.5}

\newcommand{\ApgemCMUAllCombinedRone}{3.8}
\newcommand{\ApgemCMUAllCombinedRfive}{11.9}
\newcommand{\ApgemCMUAllCombinedRten}{62.9}

\newcommand{\SuperglueCMUAllCombinedRone}{9.5}
\newcommand{\SuperglueCMUAllCombinedRfive}{30.7}
\newcommand{\SuperglueCMUAllCombinedRten}{\textbf{96.7}}

\newcommand{\OursCMUAllCombinedRone}{10.0}
\newcommand{\OursCMUAllCombinedRfive}{31.3}
\newcommand{\OursCMUAllCombinedRten}{94.5}

\newcommand{\OursThreeCMUAllCombinedRone}{\textbf{11.8}}
\newcommand{\OursThreeCMUAllCombinedRfive}{\textbf{36.2}}
\newcommand{\OursThreeCMUAllCombinedRten}{96.2}

\newcommand{\OursAppearanceCMUAllCombinedRone}{}
\newcommand{\OursAppearanceCMUAllCombinedRfive}{}
\newcommand{\OursAppearanceCMUAllCombinedRten}{}

\newcommand{\OursRansacOnlyCMUAllCombinedRone}{}
\newcommand{\OursRansacOnlyCMUAllCombinedRfive}{}
\newcommand{\OursRansacOnlyCMUAllCombinedRten}{}

\newcommand{\OursSpatialApproxCMUAllCombinedRone}{}
\newcommand{\OursSpatialApproxCMUAllCombinedRfive}{}
\newcommand{\OursSpatialApproxCMUAllCombinedRten}{}

%% file: 6-ablation-table.tex
\newcount\columncountablation
\columncountablation = 19

\input{6-ablation-table-input}

\begin{table*}[!t]
  \scriptsize
  \setlength\tabcolsep{0.1cm}
  \renewcommand{\arraystretch}{1.2}
  \centering
  \caption{Ablation study}%
  \resizebox{0.99\linewidth}{!}{
    \begin{tabular}{c|ccc||ccc||ccc||ccc||ccc||ccc}
\multirow{2}{*}{Method} & \multicolumn{3}{c||}{\textbf{Nordland}}& \multicolumn{3}{c||}{\textbf{Mapillary (Val. set)}}& \multicolumn{3}{c||}{\textbf{Pittsburgh 30k}}& \multicolumn{3}{c||}{\textbf{Tokyo 24/7}}& \multicolumn{3}{c||}{\textbf{RobotCar Seasons v2}}& \multicolumn{3}{c}{\textbf{Extended CMU Seasons}} \\
\cline{2-\columncountablation}
& R@1 & R@5 & R@10 & R@1 & R@5 & R@10 & R@1 & R@5 & R@10 & R@1 & R@5 & R@10 & .25m/2\textdegree & .5m/5\textdegree & 5.0m/10\textdegree & .25m/2\textdegree & .5m/5\textdegree & 5.0m/10\textdegree \\
\cline{1-\columncountablation}
\noalign{\vskip\doublerulesep\vskip-\arrayrulewidth}
\cline{1-\columncountablation}
Ours (Single-Spatial-Patch-NetVLAD) & \SpatialPatchNordlandRone & \SpatialPatchNordlandRfive & \SpatialPatchNordlandRten & \SpatialPatchMapillaryValRone & \SpatialPatchMapillaryValRfive & \SpatialPatchMapillaryValRten & \SpatialPatchPittsburghRone & \SpatialPatchPittsburghRfive & \SpatialPatchPittsburghRten & \SpatialPatchTokyoTFSRone & \SpatialPatchTokyoTFSRfive & \SpatialPatchTokyoTFSRten & \SpatialPatchRSAllCombinedRone & \SpatialPatchRSAllCombinedRfive & \SpatialPatchRSAllCombinedRten & \SpatialPatchCMUAllCombinedRone & \SpatialPatchCMUAllCombinedRfive & \SpatialPatchCMUAllCombinedRten \\
Ours (Single-RANSAC-Patch-NetVLAD) & \SinglePatchNordlandRone & \SinglePatchNordlandRfive & \SinglePatchNordlandRten & \SinglePatchMapillaryValRone & \SinglePatchMapillaryValRfive & \SinglePatchMapillaryValRten & \SinglePatchPittsburghRone & \SinglePatchPittsburghRfive & \SinglePatchPittsburghRten & \SinglePatchTokyoTFSRone & \SinglePatchTokyoTFSRfive & \SinglePatchTokyoTFSRten & \SinglePatchRSAllCombinedRone & \SinglePatchRSAllCombinedRfive & \SinglePatchRSAllCombinedRten & \SinglePatchCMUAllCombinedRone & \SinglePatchCMUAllCombinedRfive & \SinglePatchCMUAllCombinedRten \\
Ours (Multi-Spatial-Patch-NetVLAD) & \MultiSpatialPatchNordlandRone & \MultiSpatialPatchNordlandRfive & \MultiSpatialPatchNordlandRten & \MultiSpatialPatchMapillaryValRone & \MultiSpatialPatchMapillaryValRfive & \MultiSpatialPatchMapillaryValRten & \MultiSpatialPatchPittsburghRone & \MultiSpatialPatchPittsburghRfive & \MultiSpatialPatchPittsburghRten & \MultiSpatialPatchTokyoTFSRone & \MultiSpatialPatchTokyoTFSRfive & \MultiSpatialPatchTokyoTFSRten & \MultiSpatialPatchRSAllCombinedRone & \MultiSpatialPatchRSAllCombinedRfive & \MultiSpatialPatchRSAllCombinedRten & \MultiSpatialPatchCMUAllCombinedRone & \MultiSpatialPatchCMUAllCombinedRfive & \MultiSpatialPatchCMUAllCombinedRten \\\cline{1-\columncountablation}
\textbf{Ours (Multi-RANSAC-Patch-NetVLAD)} & \OursNordlandThreeRone & \OursNordlandThreeRfive & \OursNordlandThreeRten & \OursMapillaryValThreeRone & \OursMapillaryValThreeRfive & \OursAblMapillaryValThreeRten & \OursPittsburghThreeRone & \OursAblPittsburghThreeRfive & \OursAblPittsburghThreeRten & \OursAblTokyoTFSThreeRone & \OursAblTokyoTFSThreeRfive & \OursTokyoTFSThreeRten & \OursThreeRSAllCombinedRone & \OursThreeRSAllCombinedRfive & \OursThreeRSAllCombinedRten & \OursThreeCMUAllCombinedRone & \OursThreeCMUAllCombinedRfive & \OursThreeCMUAllCombinedRten \\
\end{tabular}%
    }
  \label{tab:results_ablation}%
  \vspace*{-0.2cm}
\end{table*}%

%% file: 6-ablation-table-input.tex
\newcommand{\SinglePatchNordlandRone}{\OursNordlandRone}
\newcommand{\SinglePatchNordlandRfive}{\OursNordlandRfive}
\newcommand{\SinglePatchNordlandRten}{\OursNordlandRten}

\newcommand{\SinglePatchMapillaryValRone}{\OursMapillaryValRone}
\newcommand{\SinglePatchMapillaryValRfive}{\OursMapillaryValRfive}
\newcommand{\SinglePatchMapillaryValRten}{\OursMapillaryValRten}

\newcommand{\SinglePatchPittsburghRone}{\OursRansacOnlyPittsburghRone}
\newcommand{\SinglePatchPittsburghRfive}{\OursRansacOnlyPittsburghRfive}
\newcommand{\SinglePatchPittsburghRten}{\OursRansacOnlyPittsburghRten}

\newcommand{\SinglePatchTokyoTFSRone}{\OursRansacOnlyTokyoTFSRone}
\newcommand{\SinglePatchTokyoTFSRfive}{\OursRansacOnlyTokyoTFSRfive}
\newcommand{\SinglePatchTokyoTFSRten}{\OursRansacOnlyTokyoTFSRten}

\newcommand{\SinglePatchRSAllCombinedRone}{\OursRSAllCombinedRone}
\newcommand{\SinglePatchRSAllCombinedRfive}{\OursRSAllCombinedRfive}
\newcommand{\SinglePatchRSAllCombinedRten}{\OursRSAllCombinedRten}

\newcommand{\SinglePatchCMUAllCombinedRone}{\OursCMUAllCombinedRone}
\newcommand{\SinglePatchCMUAllCombinedRfive}{\OursCMUAllCombinedRfive}
\newcommand{\SinglePatchCMUAllCombinedRten}{\OursCMUAllCombinedRten}

\newcommand{\SpatialPatchPittsburghRone}{88.0}
\newcommand{\SpatialPatchPittsburghRfive}{94.0}
\newcommand{\SpatialPatchPittsburghRten}{95.6}

\newcommand{\SpatialPatchNordlandRone}{42.9}
\newcommand{\SpatialPatchNordlandRfive}{49.2}
\newcommand{\SpatialPatchNordlandRten}{51.6}

\newcommand{\SpatialPatchMapillaryValRone}{77.2}
\newcommand{\SpatialPatchMapillaryValRfive}{85.4}
\newcommand{\SpatialPatchMapillaryValRten}{87.3}

\newcommand{\SpatialPatchTokyoTFSRone}{78.1}
\newcommand{\SpatialPatchTokyoTFSRfive}{83.8}
\newcommand{\SpatialPatchTokyoTFSRten}{87.0}

\newcommand{\SpatialPatchRSAllCombinedRone}{8.7}
\newcommand{\SpatialPatchRSAllCombinedRfive}{32.4}
\newcommand{\SpatialPatchRSAllCombinedRten}{88.4}

\newcommand{\SpatialPatchCMUAllCombinedRone}{10.0}
\newcommand{\SpatialPatchCMUAllCombinedRfive}{31.5}
\newcommand{\SpatialPatchCMUAllCombinedRten}{95.2}

\newcommand{\MultiSpatialPatchNordlandRone}{44.5}
\newcommand{\MultiSpatialPatchNordlandRfive}{50.1}
\newcommand{\MultiSpatialPatchNordlandRten}{52.0}

\newcommand{\MultiSpatialPatchMapillaryValRone}{78.2}
\newcommand{\MultiSpatialPatchMapillaryValRfive}{85.3}
\newcommand{\MultiSpatialPatchMapillaryValRten}{86.9}

\newcommand{\MultiSpatialPatchPittsburghRone}{88.6} \newcommand{\MultiSpatialPatchPittsburghRfive}{\textbf{94.5}} \newcommand{\MultiSpatialPatchPittsburghRten}{95.8}

\newcommand{\MultiSpatialPatchTokyoTFSRone}{81.9} \newcommand{\MultiSpatialPatchTokyoTFSRfive}{85.7} \newcommand{\MultiSpatialPatchTokyoTFSRten}{87.9}

\newcommand{\MultiSpatialPatchRSAllCombinedRone}{9.4}
\newcommand{\MultiSpatialPatchRSAllCombinedRfive}{33.9}
\newcommand{\MultiSpatialPatchRSAllCombinedRten}{89.3}

\newcommand{\MultiSpatialPatchCMUAllCombinedRone}{11.1}
\newcommand{\MultiSpatialPatchCMUAllCombinedRfive}{34.5}
\newcommand{\MultiSpatialPatchCMUAllCombinedRten}{\textbf{96.3}}

\newcommand{\OursAblMapillaryValThreeRten}{87.7}
\newcommand{\OursAblPittsburghThreeRfive}{\textbf{94.5}}
\newcommand{\OursAblPittsburghThreeRten}{\textbf{95.9}}
\newcommand{\OursAblTokyoTFSThreeRone}{\textbf{86.0}}
\newcommand{\OursAblTokyoTFSThreeRfive}{\textbf{88.6}}

%% file: 7-conclusions.tex
\section{Discussion and Conclusion}
\label{sec:discussion}
In this work we have proposed a novel \emph{locally-global} feature descriptor, which uses global descriptor techniques to further improve the appearance robustness of local descriptors. Unlike prior keypoint-based local feature descriptors \cite{lowe1999object, detone2018superpoint}, our approach considers all the visual content within a larger \emph{patch} of the image, using techniques that facilitate further performance improvements through an efficient multi-scale fusion of patches. Our proposed Patch-NetVLAD's average performance across key benchmarks is superior by 17.5\% over the original NetVLAD, and by 3.1\% (\emph{absolute} recall increase) over the state-of-the-art SuperPoint and SuperGlue-enabled VPR pipeline. Our experiments reveal an inherent benefit to fusing multiple patch sizes simultaneously, where the fused recall is greater than any single patch size recall%
, and provide a means by which to do so with minimal computational penalty compared to single scale techniques.

While this demonstration of Patch-NetVLAD occurred in a place recognition context, further applications and extensions are possible. 
One avenue for future work is the following: while we match Patch-NetVLAD features using mutual nearest neighbors with subsequent spatial verification using RANSAC, recent deep learned matchers~\cite{sarlin20superglue,zhang2019oanet} could further improve the global re-localization performance of the algorithm. 
Although our method is by no means biologically inspired, it is worth noting that the brain processes visual information over multiple receptive fields~\cite{Hubel1977}. As a result, another potentially promising direction for future research is to explore and draw inspiration from how the task of visual place recognition, rather than the more commonly studied object or face recognition tasks, is achieved in the brain. Finally, another line of work could consider the correlation between the learned VLAD clustering and semantic classes (\eg car, pedestrian, building), in order to identify and remove patches that contain dynamic objects.%

\iftoggle{cvprfinal}{\footnotesize \textbf{Acknowledgements:} We would like to thank Gustavo Carneiro and Niko Suenderhauf for their valuable comments in preparing this paper. This work received funding from the Australian Government, via grant AUSMURIB000001 associated with ONR MURI grant N00014-19-1-2571. The authors acknowledge continued support from the Queensland University of Technology (QUT) through the Centre for Robotics.
}{}

%% file: 0-additional-experimental-results.tex
\section*{Overview}
The Supplementary Material is structured as follows. In Section~\ref{sec:resultscondition}, we show results obtained on the RobotCar Seasons v2 and Extended CMU Seasons datasets split by query condition. Section~\ref{sec:quantitative} contains additional quantitative results on the Pittsburgh 30k and Tokyo 24/7 datasets, as well as additional results on the computation time split across the feature extraction and feature matching processes. Section~\ref{sec:ablation} contains a variety of additional ablation studies, some of them further demonstrating the robustness of our method, while others detail experiments that might logically be expected and were conducted, but that were not fruitful. Section~\ref{sec:qualitative} contains various qualitative results, showcasing both challenging success cases of Patch-NetVLAD and some failure cases. This section also contains examples of incorrect dataset annotations, where Patch-NetVLAD actually found the correct match but this match was not within the ground-truth matches due to errors in the ground-truth. Finally, in Section~\ref{sec:datasets} we describe in detail the six key benchmark datasets on which we evaluate Patch-NetVLAD.

\section{Results Split by Condition on RobotCar Seasons v2 and Extended CMU Seasons}
\label{sec:resultscondition}

\textbf{RobotCar Seasons v2:} 
Suppl.~Table~\ref{tab:results_robotcar} contains results obtained on the RobotCar Seasons v2 dataset split by query condition. The results for RobotCar Seasons v2 stated in Tables 1 and 2 of the main paper are summary statistics, where the different conditions are weighted by the number of images contained within each condition.

Patch-NetVLAD outperforms SuperGlue~\cite{sarlin20superglue} by 1.3\% absolute recall on the tightest error thresholds (.25m translational error and 2 degrees orientation error) when considering the summary statistic. There are some conditions where SuperGlue has a slight performance advantage for the looser error thresholds, in particular the night traverses. As stated in the Conclusions section of the main paper, it would be interesting to train a neural network-based feature matcher similar to SuperGlue that uses our proposed Patch-NetVLAD features instead of the original SuperPoint~\cite{detone2018superpoint} features. This approach would likely yield more robust matching than a standard mutual nearest neighbors matching technique, which combined with outlier rejection will likely yield a significant performance improvement.

\input{0-results-robotcar}

\textbf{Extended CMU Seasons:} 
In Suppl.~Table~\ref{tab:results_cmu} we similarly show detailed results for the Extended CMU Seasons dataset, split by Urban, Suburban and Park environments. Patch-NetVLAD consistently outperforms all comparison methods, including our competitive SuperGlue baseline, on all conditions and all error thresholds by relatively large margins, with a single exception being the park condition where SuperGlue performs slightly better for the largest error threshold.

\input{0-results-cmu}

\section{Additional Quantitative Results}
\label{sec:quantitative}
\begin{figure*}[t!]
    \centering
    \includegraphics[width=0.45\linewidth]{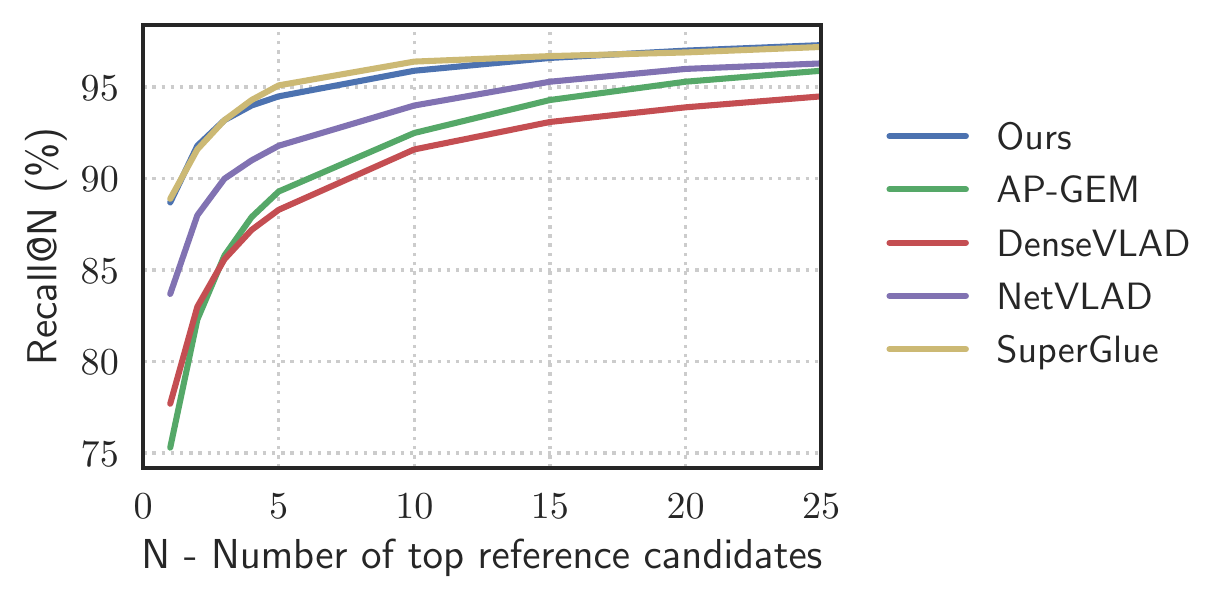}
    \hfil
    \includegraphics[width=0.45\linewidth]{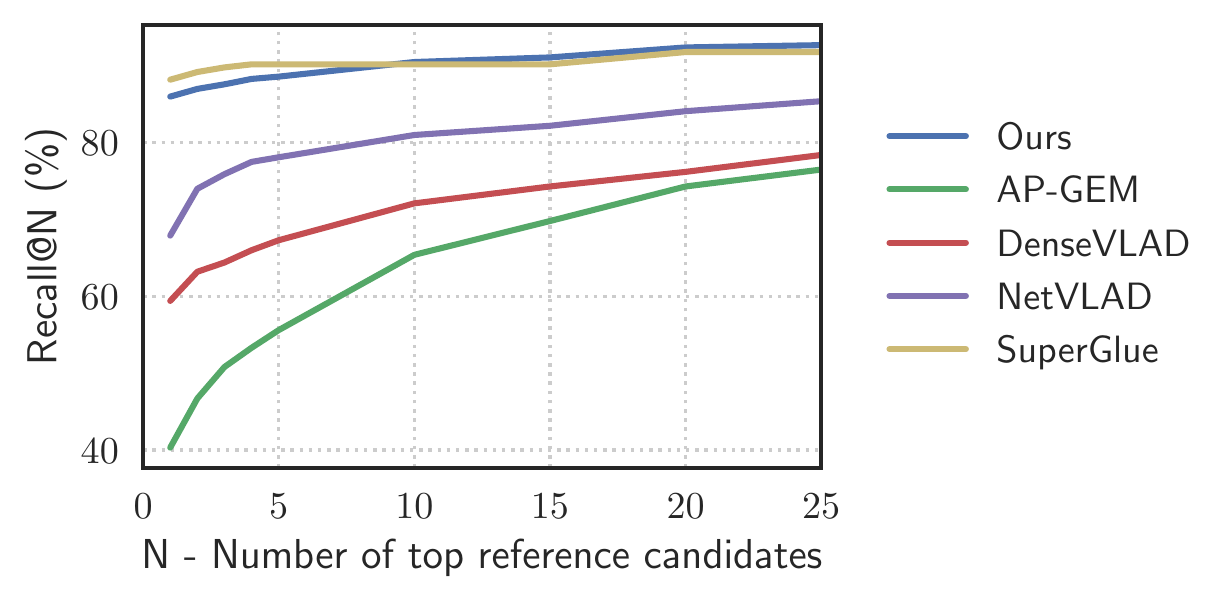}
    \caption{\textbf{Comparison with state-of-the-art.} We show the Recall$@N$ performance of Ours (Multi-RANSAC-Patch-NetVLAD) compared to AP-GEM~\cite{revaud2019learning}, DenseVLAD~\cite{Torii2018}, NetVLAD~\cite{AR2018} and SuperGlue~\cite{sarlin20superglue}, on the Pittsburgh (left) and Tokyo 24/7 (right) datasets.}
    \label{fig:recalltokyopitts}
\end{figure*}

\textbf{Additional Recall Plots:}
Fig.~3 in the main paper shows the recall@N performance on the Mapillary validation set. Similarly, Suppl.~Fig.~\ref{fig:recalltokyopitts} shows the recall@N performance for the Pittsburgh 30k and Tokyo 24/7 datasets.

\begin{figure*}[t!]
    \centering
    \includegraphics[width=0.22\textwidth]{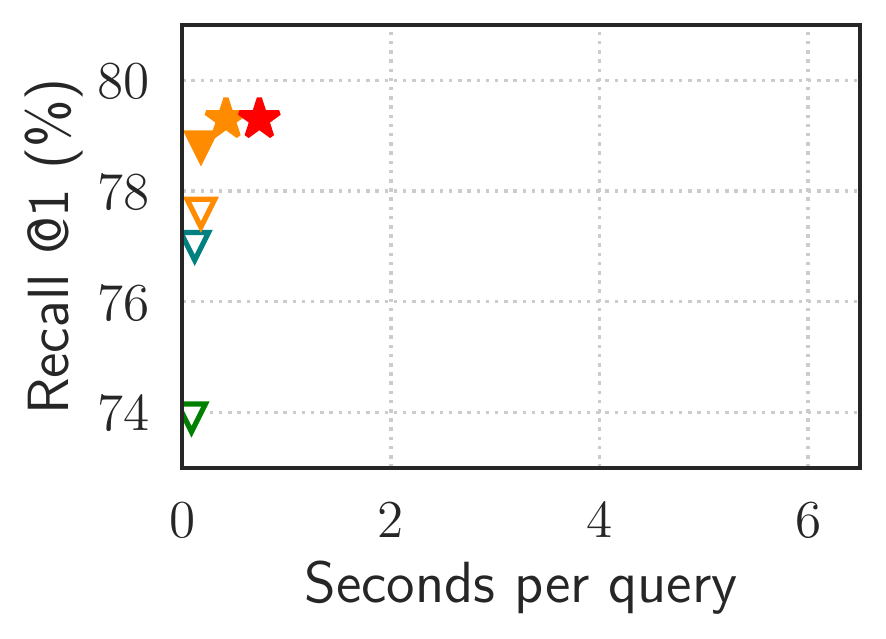}%
    \hfill
    \includegraphics[width=0.22\textwidth]{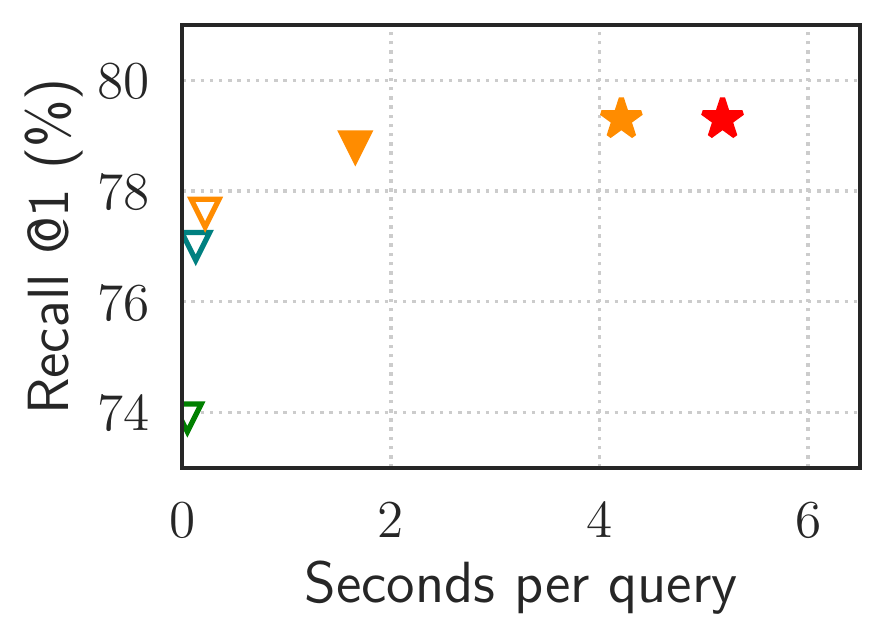}%
    \hfill
    \includegraphics[width=0.22\textwidth]{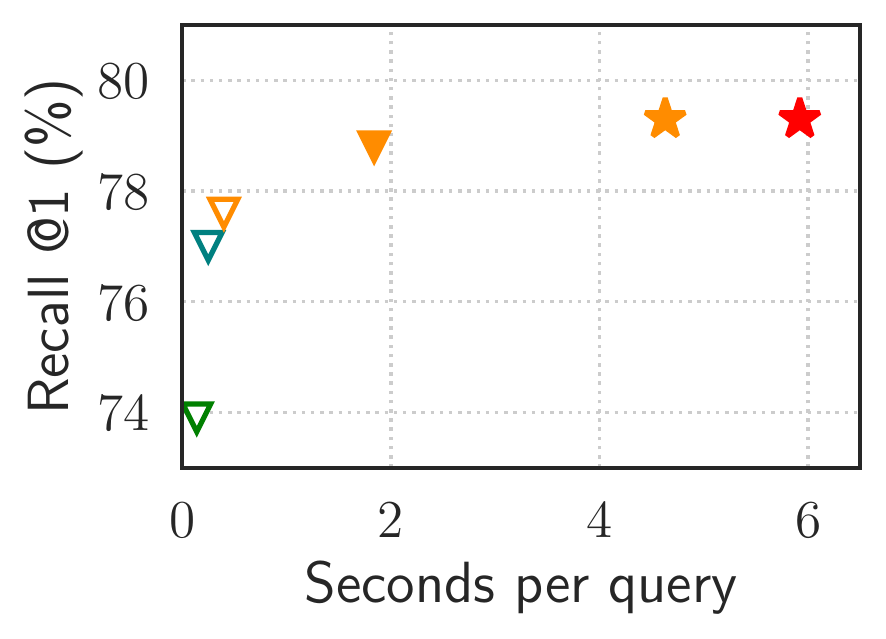}%
    \hfill
    \raisebox{0.6cm}{\includegraphics[width=0.33\textwidth]{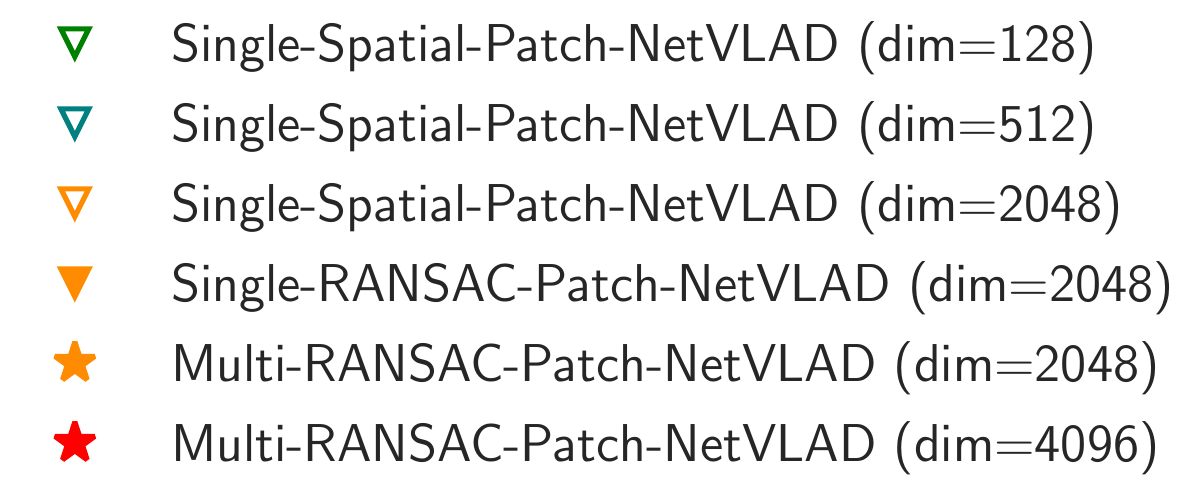}}\\
    \makebox[0.012\textwidth][c]{}
    \makebox[0.22\textwidth][c]{\small (a) Feature extraction}
    \hfill
    \makebox[0.22\textwidth][c]{\small (b) Feature matching}
    \hfill
    \makebox[0.205\textwidth][c]{\small (c) Combined}
    \hfill
    \makebox[0.315\textwidth][c]{}\\[0.2cm]
    \caption{\textbf{Computational time requirements.} The number of seconds required to process each query are shown on the $x$-axis, with the resulting R@1 shown on the $y$-axis, for the Mapillary dataset. (a) indicates the times taken for feature extraction only, while (b) shows the feature matching time. In (c) we show the combined time (as in Fig~4 in the main paper). Triangles indicate single-scale Patch-NetVLAD, while stars indicate multi-scale Patch-NetVLAD. Filled symbols are used for RANSAC matching, while hollow symbols are used for the rapid spatial verification. The color indicates varying PCA dimensions.
    }
    \label{fig:compute_ablation}
\end{figure*}

\textbf{Computational Time Requirements:}
Fig.~4 of the main paper shows the number of seconds required to process each query by a variety of our system configurations, as well as SuperGlue. The processing times presented in Fig.~4 of the main paper show the \textit{accumulated} times of feature extraction and feature matching. In Suppl.~Fig.~\ref{fig:compute_ablation} we show the compute times \textit{split} into feature extraction time only and feature matching time only; as well as the accumulated time.

\section{Further Ablation Studies}
\label{sec:ablation}

\textbf{Ablation of Multi-Scale Fusion Weights and Patch Sizes:} 
In Fig.~6 of the main paper, we demonstrated that Patch-NetVLAD is robust to the choice of particular patch sizes that are fused in our multi-scale approach. In Suppl.~Table~\ref{tab:fusionweights}, we further validate that our proposed multi-scale fusion of spatial scores across several patch sizes is robust to changes in patch size \emph{and} weightings by presenting results for the Mapillary dataset. Note that, as stated in the main paper, the set of weights used \emph{across all experiments and all datasets} was determined using a grid-search on the training set of the RobotCar Seasons v2 dataset.

While we fuse three patch sizes in the main paper, our method is not constrained to fusing any particular number of patch sizes. An investigation regarding this is shown in Suppl.~Table~\ref{tab:fusionnumber} -- there all patch sizes are fused with equal weights for simplicity. An interesting observation is that increasing the number of different patch sizes used (from three up to five) does not improve the recall performance beyond the best combination of three patch sizes. We can infer that the span of patch sizes (the difference between the smallest size and the largest size) is more important than the number of patch sizes used.

\textbf{Early Match Fusion:}
In Section 3.5 of the main paper, we describe our multi-scale fusion approach that merges the spatial scores obtained from different patch sizes. An alternative to this post-processing fusion is an early fusion where mutual nearest neighbors (Section 3.3 of the main paper) are found across patches of different scales, and a joint spatial score is calculated from all these mutual nearest neighbors.

However, we found that this early fusion approach does not work as well as the proposed post-processing fusion. Specifically, on the Mapillary validation set, we find that the early fusion results in R@1: 77.2\%, R@5: 85.3\%, and R@10: 87.3\%. This compares to R@1: 79.5\%, R@5: 86.2\% and R@10: 87.7\% using our proposed post-processing fusion.

\begin{table}[t]
  \renewcommand{\arraystretch}{1.2}
  \centering
  \caption{Ablation of Multi-Scale Fusion Weights (R@1)}
  \vspace{0.1cm}
      \footnotesize
    \begin{tabular}{ c|c|c|c } 
    Weights / Patch sizes & 2 / 5 / 8 & 2 / 4 / 6 & 2 / 6 / 10 \\
    \hline
     0.33 / 0.33 / 0.33 & \textbf{79.7} & 78.6 &78.8  \\ 
     0.2 / 0.6 / 0.2 & 79.1  & 78.5  & 78.4 \\ 
     0.45 / 0.15 / 0.4 & 79.5  & 78.9  & 78.8  \\ 
     0.45 / 0.35 / 0.2 & 78.8  & 78.1  & 79.1  \\ 
    \end{tabular}
    \label{tab:fusionweights}
\end{table}

\begin{table}[t]
  \renewcommand{\arraystretch}{1.2}
  \centering
  \caption{Ablation of the Number of Fused Patch Sizes}
  \vspace{0.1cm}
      \footnotesize
    \begin{tabular}{ c|c } 
    Patch sizes & Recall@1 \\
    \hline
     1 / 3 / 5 & 78.0 \\
     2 / 5 / 8 & \textbf{79.7}  \\ 
     3 / 5 / 7 & 79.3  \\ 
     4 / 5 / 6 & 78.2  \\ 
    \hline
     1 / 2 / 3 / 4 & 77.3  \\
     1 / 3 / 5 / 7 & 78.5  \\
     2 / 4 / 6 / 8 & 78.9  \\
    \hline
     1 / 3 / 5 / 7 / 9 & 79.5 \\
     2 / 4 / 6 / 8 / 10 & 78.8 \\
    \end{tabular}
    \label{tab:fusionnumber}
\end{table}

\textbf{Other Pooling Strategies:}
We use NetVLAD pooling to aggregate patch features into a single patch descriptor in our proposed approach. Instead of NetVLAD pooling, other pooling strategies such as max-pooling~\cite{tolias2015particular} and sum-pooling~\cite{babenko2015aggregating} have been proposed in the literature. Our spatial scoring system based on patch-based matching is in principle applicable with alternative pooling strategies. However, we found that patch-based aggregation does not perform well when applied to these pooling strategies: patch-level average pooling of VGG's Conv-5 layer (all else being equal)  improves performance from 60.8\% R@1 (vanilla NetVLAD on Mapillary dataset) to 73.6\%, which compares to 79.5\% using patch-level VLAD pooling (Patch-NetVLAD). Patch-level max-pooling similarly leads to decreased performance when compared to Patch-NetVLAD (R@1: 74.5\%). In summary, our Patch-NetVLAD description significantly outperforms those alternative pooling strategies. Further investigation will be required to gain a deeper understanding of the complementary nature of the underlying pooling strategies and our proposed patch-based aggregation.

\textbf{Patch Crops in the Image Space Instead of Feature Space:} In Patch-NetVLAD, pooling is performed from a set of patches in the \emph{feature space} of an image. One could instead perform forward passes on patch-crops in the \emph{image space}. The main problem with this approach is that processing overlapping patches is prohibitive in terms of compute and storage (as each patch needs to be separately passed through VGG). However, overlapping patches are crucial for achieving high task performance -- we found that overlapping patches are key to achieving viewpoint invariance. Therefore, performing forward passes on patch-crops in the image space is not a viable alternative to our proposed pooling of patches in the feature space.

\textbf{Matching Across Different Patch Sizes:}
In the proposed method we match patches with other patches of the same size, but there is the possibility to match between patches of different sizes. For instance, a patch of size 2x2 could find a nearest neighbor match to a patch of size 5x5. Experiments revealed that such a cross-patch-size matching leads to sub-optimal performance: R@1 reduces from 79.5\% to 78.1\%. In future works, we would like to explore other matching strategies such as a coarse-to-fine matching scheme. We would also note that, conceptually, images of different zoom levels should not be matched, as they could have been taken from different places.

\begin{figure*}[t!]
    \centering
    \includegraphics[width=0.99\textwidth]{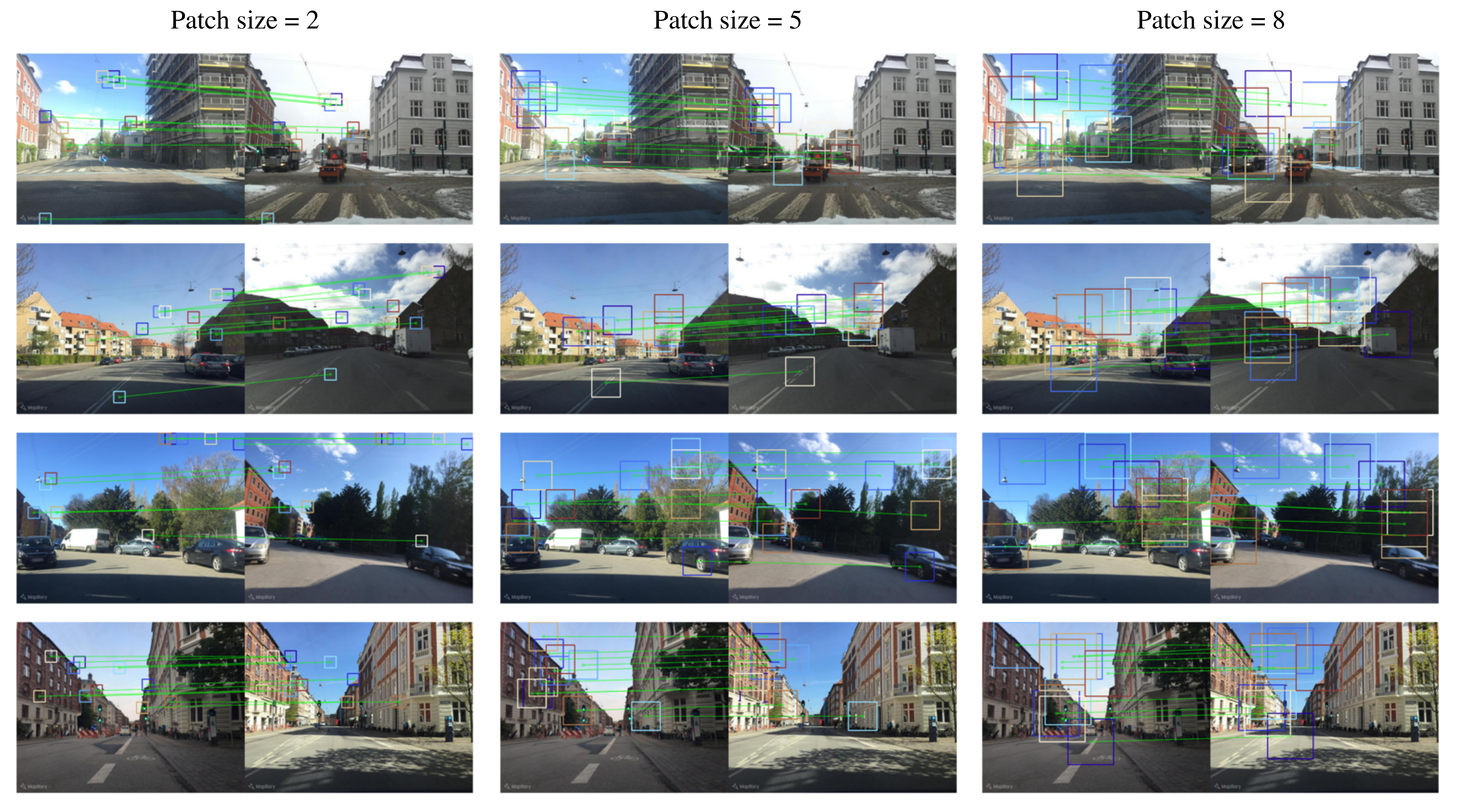}
    \caption{\textbf{Complementarity of Patch Sizes.} The three columns indicate different patch sizes, from small (\ie $d_p=2$) over medium (\ie $d_p=5$) to large (\ie $d_p=8$). It can be observed that a small patch size is able to find matches where smaller spatial context is more intuitive, for example, near boundaries between sky and buildings (first row, left column) or between sky and power lines (third row, left column). On the other hand, a larger patch size provides complementary cues by spanning over large building surfaces, enabling matching despite significant illumination variations (second row, right column). Note that the size of the squares does not reflect the receptive field sizes of the underlying features; different sizes are used for visualization purposes only.}
    \label{fig:differentpatchsizes}
\end{figure*}

\textbf{Complementarity of Patch Sizes:}
Suppl.~Fig.~\ref{fig:differentpatchsizes} shows examples of correspondences split by patch size. We randomly sampled 10 correspondences per patch size and indicate the area covered by each patch. We include examples where small/medium/large patch sizes (\ie $d_p=\{2, 5, 8\}$) result in particularly good matches, as well as one example (the bottom row) where all patch sizes work well for the same image pair but in distinct areas of the image.

\section{Additional Qualitative Results}
\label{sec:qualitative}
\begin{figure*}[t!]
    \centering
    \includegraphics[width=0.95\textwidth]{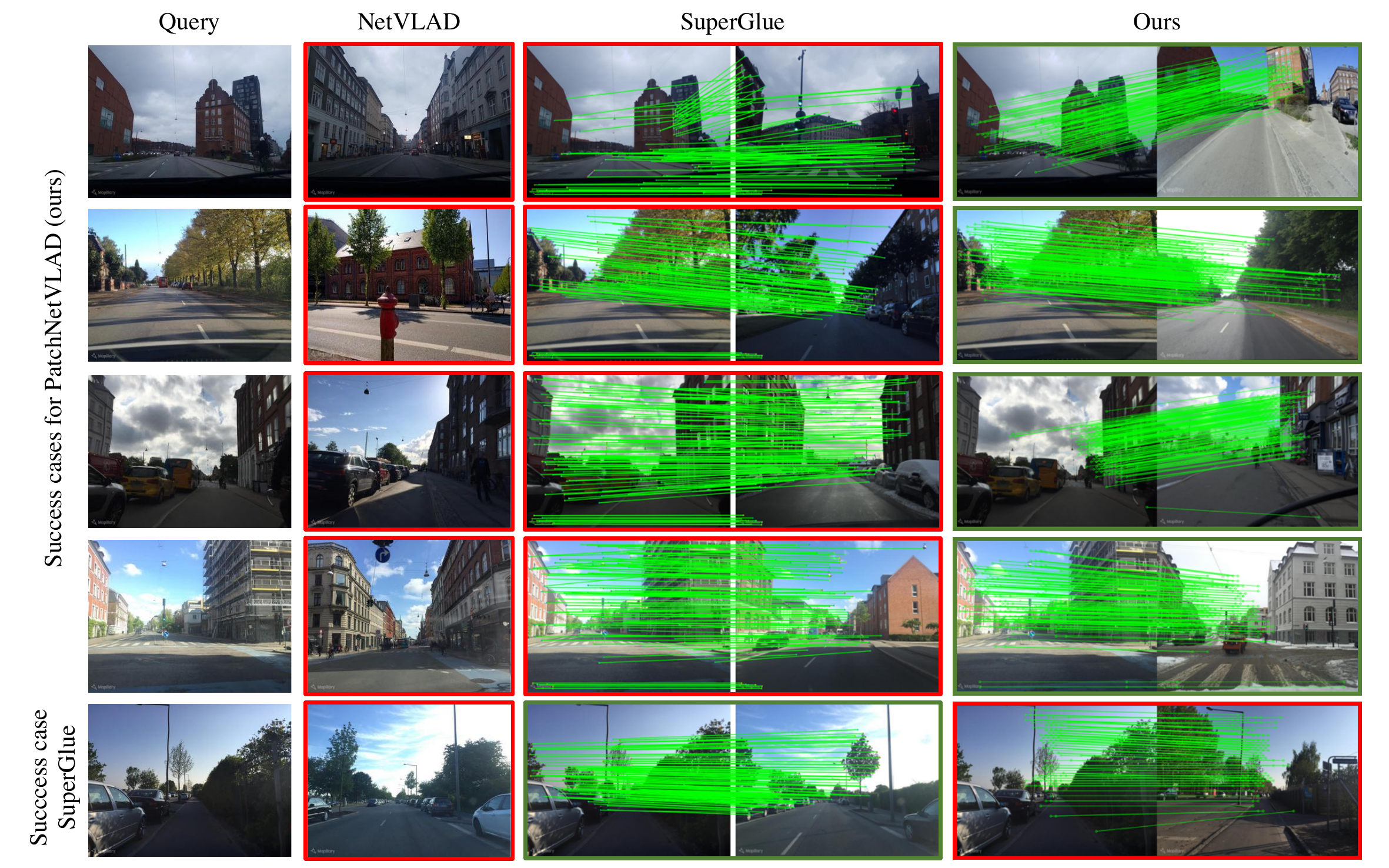}
    \caption{\textbf{Feature correspondences for the Mapillary dataset. 
    }}
    \label{fig:qualitative_mapillary}
\end{figure*}

\begin{figure*}[t!]
    \centering
    \includegraphics[width=0.95\textwidth]{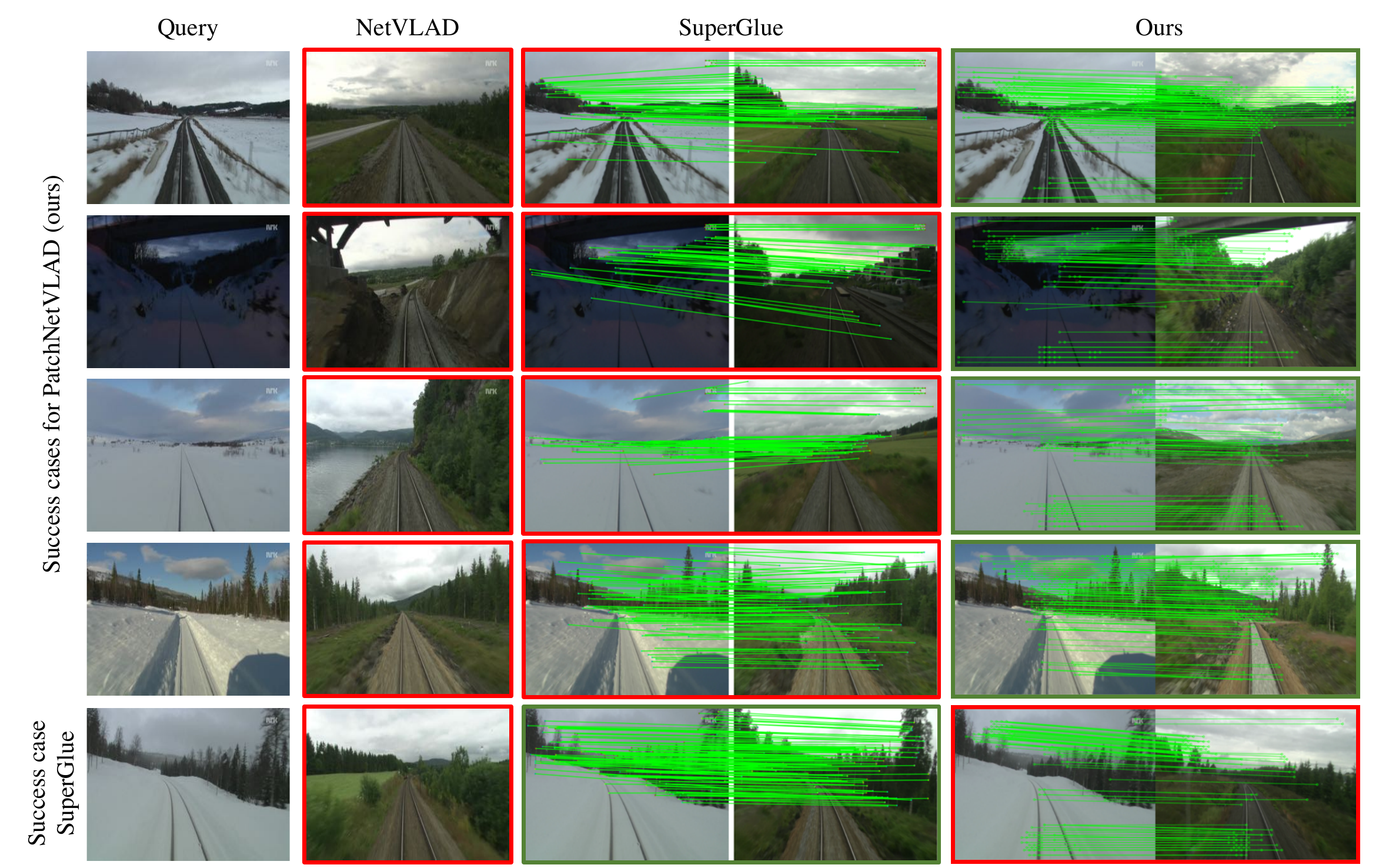}
    \caption{\textbf{Feature correspondences for the Nordland dataset.}}
    \label{fig:qualitative_nordland}
\end{figure*}

\begin{figure*}[t!]
    \centering
    \includegraphics[width=0.95\textwidth]{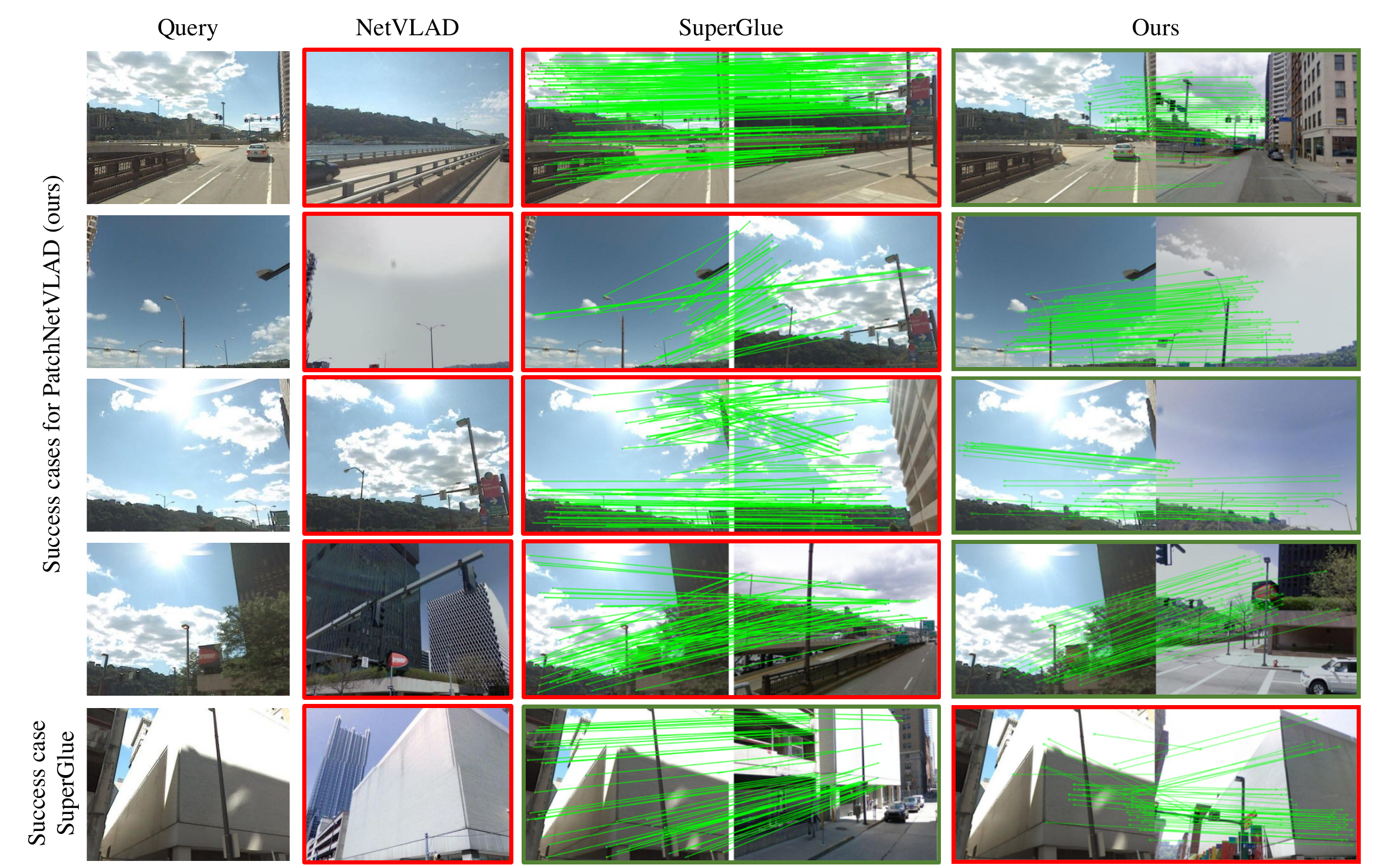}
    \caption{\textbf{Feature correspondences for the Pittsburgh dataset.}}
    \label{fig:qualitative_pittsburgh}
\end{figure*}

\begin{figure*}[t!]
    \centering
    \includegraphics[width=0.95\textwidth]{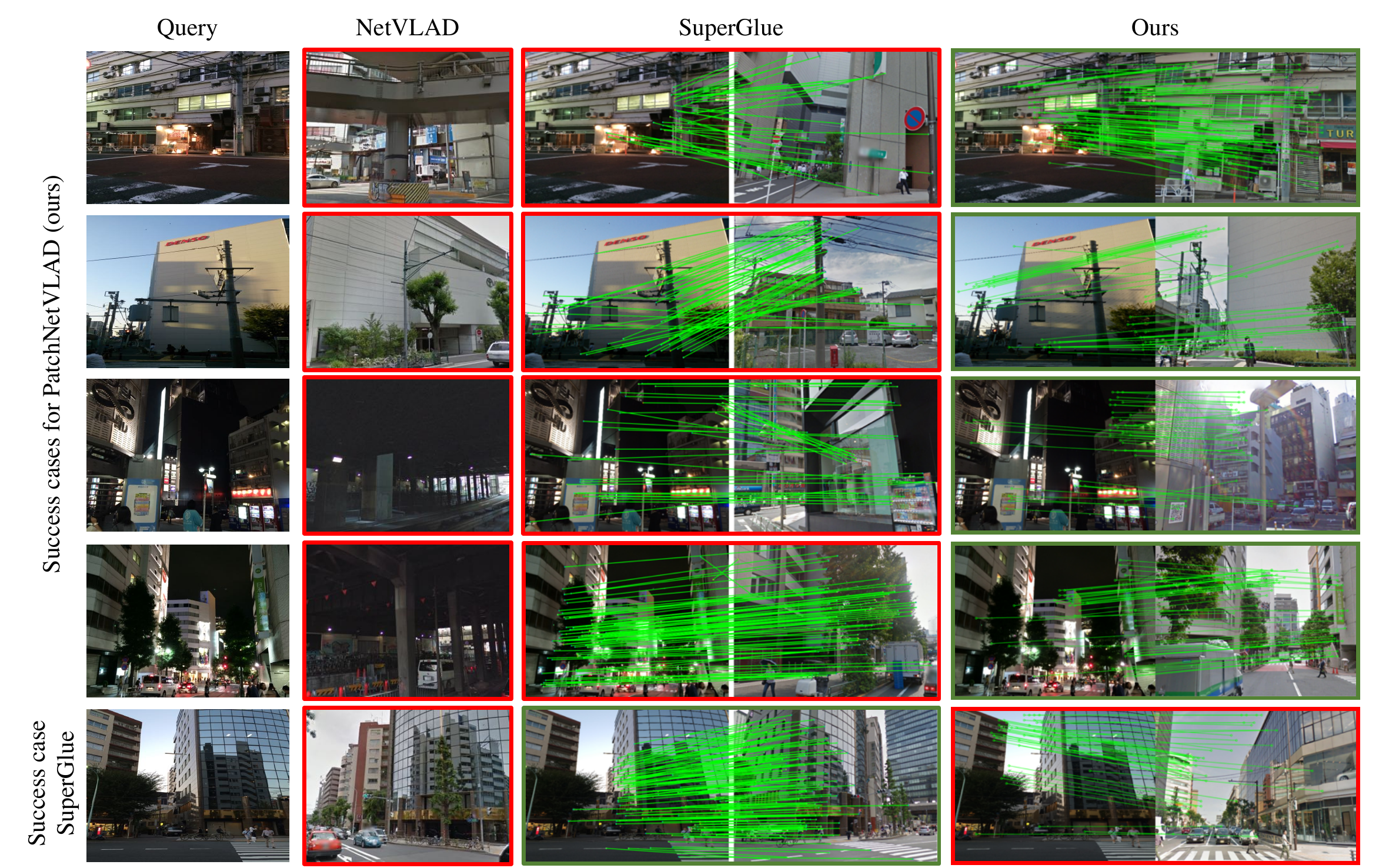}
    \caption{\textbf{Feature correspondences for the Tokyo 24/7 dataset.}}
    \label{fig:qualitative_tokyo}
\end{figure*}

Suppl.~Figs.~\ref{fig:qualitative_mapillary}, \ref{fig:qualitative_nordland}, \ref{fig:qualitative_pittsburgh} and \ref{fig:qualitative_tokyo} contain additional qualitative results on the Mapillary, Nordland, Pittsburgh and Tokyo 24/7 datasets respectively. For all these results, correct matches are represented with green borders, and incorrect matches with red borders. We show success cases of Patch-NetVLAD where all other methods failed to retrieve a correct match. Besides success cases, we also include failure cases where our proposed competitive SuperGlue baseline finds the correct match, but Patch-NetVLAD does not localize correctly. Many of these matches are challenging to recognize as the same place, even for a human observer.

These match example visualizations lead to interesting observations. For example, in Fig.~\ref{fig:qualitative_pittsburgh} on the Pittsburgh dataset, we note that a large proportion of cases where Patch-NetVLAD succeeds and Superglue fails are for images containing a large proportion of sky. We notice that SuperGlue is attempting to find correspondences between points corresponding to clouds in these images. Patch-NetVLAD, on the other hand, uses larger patch-level features which typically include clouds \textit{and} a ground level feature. Suppl. Fig~\ref{fig:differentpatchsizes} illustrates this effect by showing the corresponding patch sizes at multiple scales superimposed onto the original image.%

\begin{figure*}[t!]
    \centering
    \includegraphics[width=0.99\textwidth]{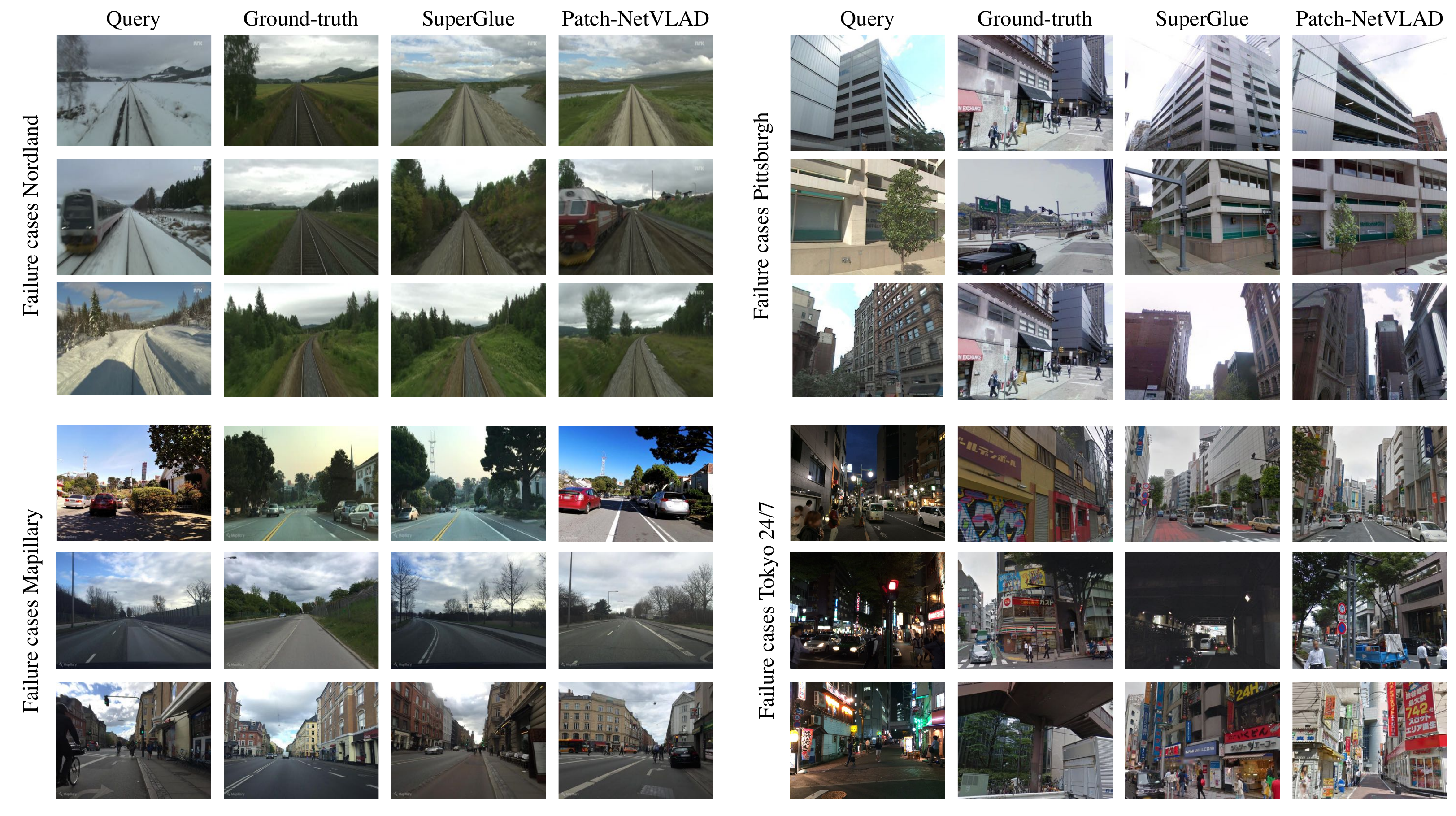}
    \caption{\textbf{Cases where \emph{all} methods fail.} We hope that these cases inform future research in VPR. Failure cases on Nordland (top left block) are mainly due to an unseen environment (for learning-based methods) and significant perceptual aliasing, \ie different places look very similar. Similarly, on the Mapillary dataset (bottom left block) both SuperGlue and Patch-NetVLAD retrieve places that have a very similar structure to the query -- consider for example the second last row where both the query and retrieved image from Patch-NetVLAD have a light pole on the left and sparse trees on right. On the Pittsburgh dataset (top right block), an additional challenge are the extreme viewpoint variations. Failures on Tokyo 24/7 (bottom right block) are mainly due to extreme viewpoint and appearance changes, as the query images are captured at night-time while reference images are captured at day-time. As mentioned in the Conclusions, we think that adding semantic information might aid Patch-NetVLAD in these extremely challenging cases.}
    \label{fig:allfail}
\end{figure*}

In Suppl.~Fig.~\ref{fig:allfail}, we showcase some examples where \emph{all} methods fail to localize correctly -- those examples may guide future research to address these open challenges.

\begin{figure*}[t!]
    \centering
    \includegraphics[width=0.99\textwidth]{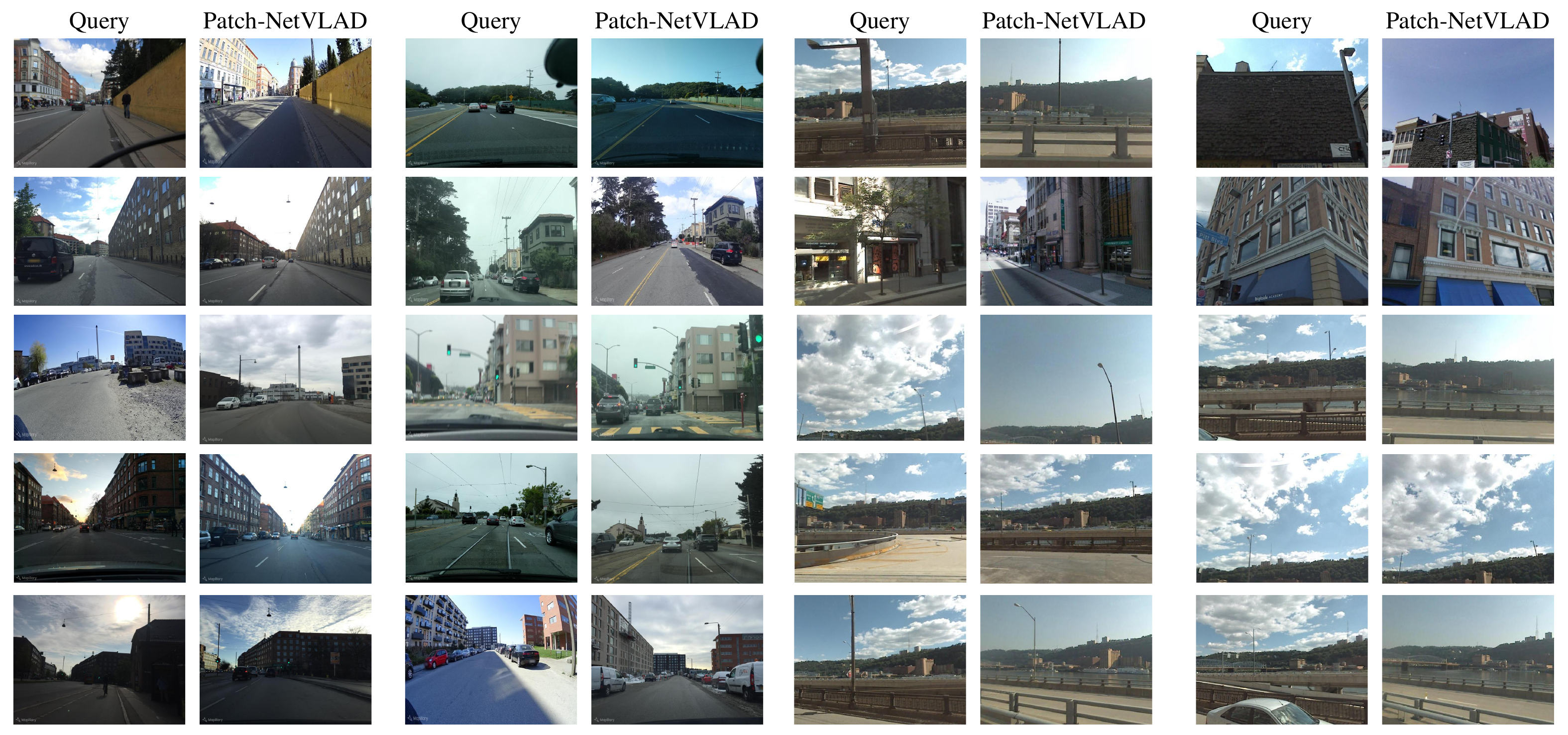}
    \caption{\textbf{Cases where Patch-NetVLAD retrieved a correct match}, but this match was deemed outside the error tolerance, suggesting either that the error tolerances are too tight (compared to what a human would consider as the same place), or the possibility of slight ground truth errors. The left columns contain examples from the Mapillary dataset, while the right columns contain examples from the Pittsburgh dataset.}
    \label{fig:wrongGT}
\end{figure*}

Finally, Suppl.~Fig.~\ref{fig:wrongGT} provides some examples of the Pittsburgh and Mapillary datasets where a manual inspection of Patch-NetVLAD's \emph{failure cases} has shown that Patch-NetVLAD \emph{actually found a correct place match}, which indicates that either the error tolerances are too tight, or that some ground-truth locations are incorrectly annotated.

%% file: 0-results-robotcar.tex
\input{0-results-robotcar-input}
\begin{table*}[!t]
  \renewcommand{\arraystretch}{1.2}
  \centering
  \caption{Performance comparison RobotCar Seasons v2}
  \resizebox{0.99\linewidth}{!}{
    \begin{tabular}{c|c|c|c|c|c|c|c|c|c}
     & \multicolumn{7}{c|}{\textbf{day conditions}} & \multicolumn{2}{c}{\textbf{night conditions}}\\
     & dawn & dusk & OC-summer & OC-winter & rain & snow & sun & night & night-rain \\
     \cline{2-10}
     \multicolumn{1}{r|}{m} & .25 / .50 / 5.0 & .25 / .50 / 5.0 & .25 / .50 / 5.0 & .25 / .50 / 5.0 & .25 / .50 / 5.0 & .25 / .50 / 5.0 & .25 / .50 / 5.0 & .25 / .50 / 5.0 & .25 / .50 / 5.0\\
     \multicolumn{1}{r|}{deg}  & 2 / 5 / 10 & 2 / 5 / 10 & 2 / 5 / 10 & 2 / 5 / 10 & 2 / 5 / 10 & 2 / 5 / 10 & 2 / 5 / 10 & 2 / 5 / 10 & 2 / 5 / 10\\
    \cline{1-10}
    \noalign{\vskip\doublerulesep\vskip-\arrayrulewidth}
    \cline{1-10}
    AP-GEM~\cite{revaud2019learning} & \ApgemRSDawnRone\ / \ApgemRSDawnRfive\ / \ApgemRSDawnRten & \ApgemRSDuskRone\ / \ApgemRSDuskRfive\ / \ApgemRSDuskRten & \ApgemRSOCSummerRone\ / \ApgemRSOCSummerRfive\ / \ApgemRSOCSummerRten & \ApgemRSOCWinterRone\ / \ApgemRSOCWinterRfive\ / \ApgemRSOCWinterRten & \ApgemRSRainRone\ / \ApgemRSRainRfive\ / \ApgemRSRainRten & \ApgemRSSnowRone\ / \ApgemRSSnowRfive\ / \ApgemRSSnowRten & \ApgemRSSunRone\ / \ApgemRSSunRfive\ / \ApgemRSSunRten & \ApgemRSNightRone\ / \ApgemRSNightRfive\ / \ApgemRSNightRten & \ApgemRSNightRainRone\ / \ApgemRSNightRainRfive\ / \ApgemRSNightRainRten\\
    DenseVLAD~\cite{Torii2018} & \DensevladRSDawnRone\ / \DensevladRSDawnRfive\ / \DensevladRSDawnRten & \DensevladRSDuskRone\ / \DensevladRSDuskRfive\ / \DensevladRSDuskRten & \DensevladRSOCSummerRone\ / \DensevladRSOCSummerRfive\ / \DensevladRSOCSummerRten & \DensevladRSOCWinterRone\ / \DensevladRSOCWinterRfive\ / \DensevladRSOCWinterRten & \DensevladRSRainRone\ / \DensevladRSRainRfive\ / \DensevladRSRainRten & \DensevladRSSnowRone\ / \DensevladRSSnowRfive\ / \DensevladRSSnowRten & \DensevladRSSunRone\ / \DensevladRSSunRfive\ / \DensevladRSSunRten & \DensevladRSNightRone\ / \DensevladRSNightRfive\ / \DensevladRSNightRten & \DensevladRSNightRainRone\ / \DensevladRSNightRainRfive\ / \DensevladRSNightRainRten\\
    NetVLAD~\cite{AR2018} & \NetvladRSDawnRone\ / \NetvladRSDawnRfive\ / \NetvladRSDawnRten & \NetvladRSDuskRone\ / \NetvladRSDuskRfive\ / \NetvladRSDuskRten & \NetvladRSOCSummerRone\ / \NetvladRSOCSummerRfive\ / \NetvladRSOCSummerRten & \NetvladRSOCWinterRone\ / \NetvladRSOCWinterRfive\ / \NetvladRSOCWinterRten & \NetvladRSRainRone\ / \NetvladRSRainRfive\ / \NetvladRSRainRten & \NetvladRSSnowRone\ / \NetvladRSSnowRfive\ / \NetvladRSSnowRten & \NetvladRSSunRone\ / \NetvladRSSunRfive\ / \NetvladRSSunRten & \NetvladRSNightRone\ / \NetvladRSNightRfive\ / \NetvladRSNightRten & \NetvladRSNightRainRone\ / \NetvladRSNightRainRfive\ / \NetvladRSNightRainRten\\
    SuperGlue~\cite{sarlin20superglue} & \SuperglueRSDawnRone\ / \SuperglueRSDawnRfive\ / \SuperglueRSDawnRten & \SuperglueRSDuskRone\ / \SuperglueRSDuskRfive\ / \SuperglueRSDuskRten & \SuperglueRSOCSummerRone\ / \SuperglueRSOCSummerRfive\ / \SuperglueRSOCSummerRten & \SuperglueRSOCWinterRone\ / \SuperglueRSOCWinterRfive\ / \SuperglueRSOCWinterRten & \SuperglueRSRainRone\ / \SuperglueRSRainRfive\ / \SuperglueRSRainRten & \SuperglueRSSnowRone\ / \SuperglueRSSnowRfive\ / \SuperglueRSSnowRten & \SuperglueRSSunRone\ / \SuperglueRSSunRfive\ / \SuperglueRSSunRten & \SuperglueRSNightRone\ / \SuperglueRSNightRfive\ / \SuperglueRSNightRten & \SuperglueRSNightRainRone\ / \SuperglueRSNightRainRfive\ / \SuperglueRSNightRainRten\\
    \cline{1-10}
    \noalign{\vskip\doublerulesep\vskip-\arrayrulewidth}
    \cline{1-10}
    \textbf{Ours} & \OursThreeRSDawnRone\ / \OursThreeRSDawnRfive\ / \OursThreeRSDawnRten & \OursThreeRSDuskRone\ / \OursThreeRSDuskRfive\ / \OursThreeRSDuskRten & \OursThreeRSOCSummerRone\ / \OursThreeRSOCSummerRfive\ / \OursThreeRSOCSummerRten & \OursThreeRSOCWinterRone\ / \OursThreeRSOCWinterRfive\ / \OursThreeRSOCWinterRten & \OursThreeRSRainRone\ / \OursThreeRSRainRfive\ / \textbf{\OursThreeRSRainRten} & \OursThreeRSSnowRone\ / \OursThreeRSSnowRfive\ / \OursThreeRSSnowRten & \OursThreeRSSunRone\ / \OursThreeRSSunRfive\ / \OursThreeRSSunRten & \OursThreeRSNightRone\ / \OursThreeRSNightRfive\ / \OursThreeRSNightRten & \OursThreeRSNightRainRone\ / \OursThreeRSNightRainRfive\ / \OursThreeRSNightRainRten
    \end{tabular}%
    }
  \label{tab:results_robotcar}%
\end{table*}%

%% file: 0-results-robotcar-input.tex
\newcommand{\NetvladRSDawnRone}{3.3}
\newcommand{\NetvladRSDawnRfive}{55.4}
\newcommand{\NetvladRSDawnRten}{72.8}

\newcommand{\DensevladRSDawnRone}{3.8}
\newcommand{\DensevladRSDawnRfive}{65.4}
\newcommand{\DensevladRSDawnRten}{81.9}

\newcommand{\ApgemRSDawnRone}{1.4}
\newcommand{\ApgemRSDawnRfive}{42.8}
\newcommand{\ApgemRSDawnRten}{61.3}

\newcommand{\SuperglueRSDawnRone}{4.3}
\newcommand{\SuperglueRSDawnRfive}{69.0}
\newcommand{\SuperglueRSDawnRten}{84.8}

\newcommand{\OursRSDawnRone}{\textbf{4.8}}
\newcommand{\OursRSDawnRfive}{70.6}
\newcommand{\OursRSDawnRten}{85.5}

\newcommand{\OursThreeRSDawnRone}{\textbf{4.8}}
\newcommand{\OursThreeRSDawnRfive}{\textbf{72.5}}
\newcommand{\OursThreeRSDawnRten}{\textbf{86.2}}

\newcommand{\OursAppearanceRSDawnRone}{}
\newcommand{\OursAppearanceRSDawnRfive}{}
\newcommand{\OursAppearanceRSDawnRten}{}

\newcommand{\OursRansacOnlyRSDawnRone}{}
\newcommand{\OursRansacOnlyRSDawnRfive}{}
\newcommand{\OursRansacOnlyRSDawnRten}{}

\newcommand{\OursSpatialApproxRSDawnRone}{}
\newcommand{\OursSpatialApproxRSDawnRfive}{}
\newcommand{\OursSpatialApproxRSDawnRten}{}

\newcommand{\NetvladRSDuskRone}{10.7}
\newcommand{\NetvladRSDuskRfive}{65.2}
\newcommand{\NetvladRSDuskRten}{85.2}

\newcommand{\DensevladRSDuskRone}{\textbf{13.7}}
\newcommand{\DensevladRSDuskRfive}{68.6}
\newcommand{\DensevladRSDuskRten}{88.4}

\newcommand{\ApgemRSDuskRone}{\ 9.1}
\newcommand{\ApgemRSDuskRfive}{60.2}
\newcommand{\ApgemRSDuskRten}{83.1}

\newcommand{\SuperglueRSDuskRone}{12.7}
\newcommand{\SuperglueRSDuskRfive}{69.5}
\newcommand{\SuperglueRSDuskRten}{88.6}

\newcommand{\OursRSDuskRone}{13.0}
\newcommand{\OursRSDuskRfive}{71.7}
\newcommand{\OursRSDuskRten}{89.7}

\newcommand{\OursThreeRSDuskRone}{13.5}
\newcommand{\OursThreeRSDuskRfive}{\textbf{72.0}}
\newcommand{\OursThreeRSDuskRten}{\textbf{89.5}}

\newcommand{\OursAppearanceRSDuskRone}{}
\newcommand{\OursAppearanceRSDuskRfive}{}
\newcommand{\OursAppearanceRSDuskRten}{}

\newcommand{\OursRansacOnlyRSDuskRone}{}
\newcommand{\OursRansacOnlyRSDuskRfive}{}
\newcommand{\OursRansacOnlyRSDuskRten}{}

\newcommand{\OursSpatialApproxRSDuskRone}{}
\newcommand{\OursSpatialApproxRSDuskRfive}{}
\newcommand{\OursSpatialApproxRSDuskRten}{}

\newcommand{\NetvladRSOCSummerRone}{2.9}
\newcommand{\NetvladRSOCSummerRfive}{70.5}
\newcommand{\NetvladRSOCSummerRten}{88.2}

\newcommand{\DensevladRSOCSummerRone}{4.2}
\newcommand{\DensevladRSOCSummerRfive}{75.0}
\newcommand{\DensevladRSOCSummerRten}{90.9}

\newcommand{\ApgemRSOCSummerRone}{2.3}
\newcommand{\ApgemRSOCSummerRfive}{55.8}
\newcommand{\ApgemRSOCSummerRten}{79.4}

\newcommand{\SuperglueRSOCSummerRone}{5.0}
\newcommand{\SuperglueRSOCSummerRfive}{79.4}
\newcommand{\SuperglueRSOCSummerRten}{\textbf{95.0}}

\newcommand{\OursRSOCSummerRone}{4.8}
\newcommand{\OursRSOCSummerRfive}{80.5}
\newcommand{\OursRSOCSummerRten}{94.1}

\newcommand{\OursThreeRSOCSummerRone}{\textbf{5.3}}
\newcommand{\OursThreeRSOCSummerRfive}{\textbf{80.9}}
\newcommand{\OursThreeRSOCSummerRten}{94.5}

\newcommand{\OursAppearanceRSOCSummerRone}{}
\newcommand{\OursAppearanceRSOCSummerRfive}{}
\newcommand{\OursAppearanceRSOCSummerRten}{}

\newcommand{\OursRansacOnlyRSOCSummerRone}{}
\newcommand{\OursRansacOnlyRSOCSummerRfive}{}
\newcommand{\OursRansacOnlyRSOCSummerRten}{}

\newcommand{\OursSpatialApproxRSOCSummerRone}{}
\newcommand{\OursSpatialApproxRSOCSummerRfive}{}
\newcommand{\OursSpatialApproxRSOCSummerRten}{}

\newcommand{\NetvladRSOCWinterRone}{2.3}
\newcommand{\NetvladRSOCWinterRfive}{56.9}
\newcommand{\NetvladRSOCWinterRten}{80.5}

\newcommand{\DensevladRSOCWinterRone}{4.3}
\newcommand{\DensevladRSOCWinterRfive}{66.3}
\newcommand{\DensevladRSOCWinterRten}{86.3}

\newcommand{\ApgemRSOCWinterRone}{3.1}
\newcommand{\ApgemRSOCWinterRfive}{50.3}
\newcommand{\ApgemRSOCWinterRten}{76.1}

\newcommand{\SuperglueRSOCWinterRone}{4.5}
\newcommand{\SuperglueRSOCWinterRfive}{69.1}
\newcommand{\SuperglueRSOCWinterRten}{88.6}

\newcommand{\OursRSOCWinterRone}{5.8}
\newcommand{\OursRSOCWinterRfive}{\textbf{71.3}}
\newcommand{\OursRSOCWinterRten}{89.6}

\newcommand{\OursThreeRSOCWinterRone}{\textbf{6.3}}
\newcommand{\OursThreeRSOCWinterRfive}{\textbf{71.3}}
\newcommand{\OursThreeRSOCWinterRten}{\textbf{89.8}}

\newcommand{\OursAppearanceRSOCWinterRone}{}
\newcommand{\OursAppearanceRSOCWinterRfive}{}
\newcommand{\OursAppearanceRSOCWinterRten}{}

\newcommand{\OursRansacOnlyRSOCWinterRone}{}
\newcommand{\OursRansacOnlyRSOCWinterRfive}{}
\newcommand{\OursRansacOnlyRSOCWinterRten}{}

\newcommand{\OursSpatialApproxRSOCWinterRone}{}
\newcommand{\OursSpatialApproxRSOCWinterRfive}{}
\newcommand{\OursSpatialApproxRSOCWinterRten}{}

\newcommand{\NetvladRSRainRone}{4.2}
\newcommand{\NetvladRSRainRfive}{71.9}
\newcommand{\NetvladRSRainRten}{88.9}

\newcommand{\DensevladRSRainRone}{5.4}
\newcommand{\DensevladRSRainRfive}{76.8}
\newcommand{\DensevladRSRainRten}{92.1}

\newcommand{\ApgemRSRainRone}{4.4}
\newcommand{\ApgemRSRainRfive}{67.2}
\newcommand{\ApgemRSRainRten}{87.0}

\newcommand{\SuperglueRSRainRone}{\textbf{5.9}}
\newcommand{\SuperglueRSRainRfive}{76.4}
\newcommand{\SuperglueRSRainRten}{91.8}

\newcommand{\OursRSRainRone}{\textbf{5.9}}
\newcommand{\OursRSRainRfive}{79.0}
\newcommand{\OursRSRainRten}{\textbf{92.3}}

\newcommand{\OursThreeRSRainRone}{\textbf{5.9}}
\newcommand{\OursThreeRSRainRfive}{\textbf{79.3}}
\newcommand{\OursThreeRSRainRten}{92.1}

\newcommand{\OursAppearanceRSRainRone}{}
\newcommand{\OursAppearanceRSRainRfive}{}
\newcommand{\OursAppearanceRSRainRten}{}

\newcommand{\OursRansacOnlyRSRainRone}{}
\newcommand{\OursRansacOnlyRSRainRfive}{}
\newcommand{\OursRansacOnlyRSRainRten}{}

\newcommand{\OursSpatialApproxRSRainRone}{}
\newcommand{\OursSpatialApproxRSRainRfive}{}
\newcommand{\OursSpatialApproxRSRainRten}{}

\newcommand{\NetvladRSSnowRone}{4.5}
\newcommand{\NetvladRSSnowRfive}{63.7}
\newcommand{\NetvladRSSnowRten}{80.1}

\newcommand{\DensevladRSSnowRone}{7.3}
\newcommand{\DensevladRSSnowRfive}{72.4}
\newcommand{\DensevladRSSnowRten}{85.1}

\newcommand{\ApgemRSSnowRone}{2.1}
\newcommand{\ApgemRSSnowRfive}{51.9}
\newcommand{\ApgemRSSnowRten}{70.0}

\newcommand{\SuperglueRSSnowRone}{7.0}
\newcommand{\SuperglueRSSnowRfive}{73.2}
\newcommand{\SuperglueRSSnowRten}{87.2}

\newcommand{\OursRSSnowRone}{7.4}
\newcommand{\OursRSSnowRfive}{74.2}
\newcommand{\OursRSSnowRten}{87.0}

\newcommand{\OursThreeRSSnowRone}{\textbf{7.8}}
\newcommand{\OursThreeRSSnowRfive}{\textbf{75.9}}
\newcommand{\OursThreeRSSnowRten}{\textbf{87.9}}

\newcommand{\OursAppearanceRSSnowRone}{}
\newcommand{\OursAppearanceRSSnowRfive}{}
\newcommand{\OursAppearanceRSSnowRten}{}

\newcommand{\OursRansacOnlyRSSnowRone}{}
\newcommand{\OursRansacOnlyRSSnowRfive}{}
\newcommand{\OursRansacOnlyRSSnowRten}{}

\newcommand{\OursSpatialApproxRSSnowRone}{}
\newcommand{\OursSpatialApproxRSSnowRfive}{}
\newcommand{\OursSpatialApproxRSSnowRten}{}

\newcommand{\NetvladRSSunRone}{2.2}
\newcommand{\NetvladRSSunRfive}{48.2}
\newcommand{\NetvladRSSunRten}{70.3}

\newcommand{\DensevladRSSunRone}{3.0}
\newcommand{\DensevladRSSunRfive}{55.5}
\newcommand{\DensevladRSSunRten}{71.6}

\newcommand{\ApgemRSSunRone}{0.8}
\newcommand{\ApgemRSSunRfive}{29.3}
\newcommand{\ApgemRSSunRten}{53.1}

\newcommand{\SuperglueRSSunRone}{3.3}
\newcommand{\SuperglueRSSunRfive}{66.5}
\newcommand{\SuperglueRSSunRten}{\textbf{83.9}}

\newcommand{\OursRSSunRone}{4.0}
\newcommand{\OursRSSunRfive}{66.0}
\newcommand{\OursRSSunRten}{82.3}

\newcommand{\OursThreeRSSunRone}{\textbf{4.8}}
\newcommand{\OursThreeRSSunRfive}{\textbf{67.3}}
\newcommand{\OursThreeRSSunRten}{83.4}

\newcommand{\OursAppearanceRSSunRone}{}
\newcommand{\OursAppearanceRSSunRfive}{}
\newcommand{\OursAppearanceRSSunRten}{}

\newcommand{\OursRansacOnlyRSSunRone}{}
\newcommand{\OursRansacOnlyRSSunRfive}{}
\newcommand{\OursRansacOnlyRSSunRten}{}

\newcommand{\OursSpatialApproxRSSunRone}{}
\newcommand{\OursSpatialApproxRSSunRfive}{}
\newcommand{\OursSpatialApproxRSSunRten}{}

\newcommand{\NetvladRSNightRone}{0.2}
\newcommand{\NetvladRSNightRfive}{2.5}
\newcommand{\NetvladRSNightRten}{5.4\ }

\newcommand{\DensevladRSNightRone}{0.5}
\newcommand{\DensevladRSNightRfive}{9.8}
\newcommand{\DensevladRSNightRten}{17.3}

\newcommand{\ApgemRSNightRone}{0.2}
\newcommand{\ApgemRSNightRfive}{1.7}
\newcommand{\ApgemRSNightRten}{3.4\ }

\newcommand{\SuperglueRSNightRone}{\textbf{0.5}}
\newcommand{\SuperglueRSNightRfive}{\textbf{13.4}}
\newcommand{\SuperglueRSNightRten}{\textbf{27.9}}

\newcommand{\OursRSNightRone}{0.3}
\newcommand{\OursRSNightRfive}{10.0}
\newcommand{\OursRSNightRten}{21.7}

\newcommand{\OursThreeRSNightRone}{\textbf{0.5}}
\newcommand{\OursThreeRSNightRfive}{12.4}
\newcommand{\OursThreeRSNightRten}{24.9}

\newcommand{\OursAppearanceRSNightRone}{}
\newcommand{\OursAppearanceRSNightRfive}{}
\newcommand{\OursAppearanceRSNightRten}{}

\newcommand{\OursRansacOnlyRSNightRone}{}
\newcommand{\OursRansacOnlyRSNightRfive}{}
\newcommand{\OursRansacOnlyRSNightRten}{}

\newcommand{\OursSpatialApproxRSNightRone}{}
\newcommand{\OursSpatialApproxRSNightRfive}{}
\newcommand{\OursSpatialApproxRSNightRten}{}

\newcommand{\NetvladRSNightRainRone}{0.1\ }
\newcommand{\NetvladRSNightRainRfive}{2.4\ }
\newcommand{\NetvladRSNightRainRten}{5.5\ }

\newcommand{\DensevladRSNightRainRone}{0.6}
\newcommand{\DensevladRSNightRainRfive}{17.9}
\newcommand{\DensevladRSNightRainRten}{29.6}

\newcommand{\ApgemRSNightRainRone}{0.0\,}
\newcommand{\ApgemRSNightRainRfive}{\,1.6\ }
\newcommand{\ApgemRSNightRainRten}{7.3\ }

\newcommand{\SuperglueRSNightRainRone}{0.9}
\newcommand{\SuperglueRSNightRainRfive}{\textbf{20.5}}
\newcommand{\SuperglueRSNightRainRten}{\textbf{31.8}}

\newcommand{\OursRSNightRainRone}{0.4}
\newcommand{\OursRSNightRainRfive}{17.0}
\newcommand{\OursRSNightRainRten}{27.7}

\newcommand{\OursThreeRSNightRainRone}{\textbf{1.0}}
\newcommand{\OursThreeRSNightRainRfive}{19.0}
\newcommand{\OursThreeRSNightRainRten}{30.8}

\newcommand{\OursAppearanceRSNightRainRone}{}
\newcommand{\OursAppearanceRSNightRainRfive}{}
\newcommand{\OursAppearanceRSNightRainRten}{}

\newcommand{\OursRansacOnlyRSNightRainRone}{}
\newcommand{\OursRansacOnlyRSNightRainRfive}{}
\newcommand{\OursRansacOnlyRSNightRainRten}{}

\newcommand{\OursSpatialApproxRSNightRainRone}{}
\newcommand{\OursSpatialApproxRSNightRainRfive}{}
\newcommand{\OursSpatialApproxRSNightRainRten}{}

%% file: 0-results-cmu.tex
\input{0-results-cmu-input}
\begin{table}[!t]
  \renewcommand{\arraystretch}{1.2}
  \centering
  \caption{Performance comparison Extended CMU Seasons}
  \resizebox{0.99\linewidth}{!}{
    \begin{tabular}{c|c|c|c}
     & Urban & Suburban & Park \\
     \cline{2-4}
     \multicolumn{1}{r|}{m} & .25 / .50 / 5.0 & .25 / .50 / 5.0 & .25 / .50 / 5.0\\
     \multicolumn{1}{r|}{deg}  & 2 / 5 / 10 & 2 / 5 / 10 & 2 / 5 / 10\\
    \cline{1-4}
    \noalign{\vskip\doublerulesep\vskip-\arrayrulewidth}
    \cline{1-4}
    AP-GEM~\cite{revaud2019learning} & \ApgemCMUUrbanRone\ / \ApgemCMUUrbanRfive\ / \ApgemCMUUrbanRten & \ApgemCMUSuburbanRone\ / \ApgemCMUSuburbanRfive\ / \ApgemCMUSuburbanRten & \ApgemCMUParkRone\ / \ApgemCMUParkRfive\ / \ApgemCMUParkRten\\
    DenseVLAD~\cite{Torii2018} & \DensevladCMUUrbanRone\ / \DensevladCMUUrbanRfive\ / \DensevladCMUUrbanRten & \DensevladCMUSuburbanRone\ / \DensevladCMUSuburbanRfive\ / \DensevladCMUSuburbanRten & \DensevladCMUParkRone\ / \DensevladCMUParkRfive\ / \DensevladCMUParkRten\\
    NetVLAD~\cite{AR2018} & \NetvladCMUUrbanRone\ / \NetvladCMUUrbanRfive\ / \NetvladCMUUrbanRten & \NetvladCMUSuburbanRone\ / \NetvladCMUSuburbanRfive\ / \NetvladCMUSuburbanRten & \NetvladCMUParkRone\ / \NetvladCMUParkRfive\ / \NetvladCMUParkRten\\
    SuperGlue~\cite{sarlin20superglue} & \SuperglueCMUUrbanRone\ / \SuperglueCMUUrbanRfive\ / \SuperglueCMUUrbanRten & \SuperglueCMUSuburbanRone\ / \SuperglueCMUSuburbanRfive\ / \SuperglueCMUSuburbanRten & \SuperglueCMUParkRone\ / \SuperglueCMUParkRfive\ / \SuperglueCMUParkRten\\
    \cline{1-4}
    \noalign{\vskip\doublerulesep\vskip-\arrayrulewidth}
    \cline{1-4}
    \textbf{Ours} & \OursThreeCMUUrbanRone\ / \OursThreeCMUUrbanRfive\ / \OursThreeCMUUrbanRten & \OursThreeCMUSuburbanRone\ / \OursThreeCMUSuburbanRfive\ / \OursThreeCMUSuburbanRten & \OursThreeCMUParkRone\ / \OursThreeCMUParkRfive\ / \OursThreeCMUParkRten
    \end{tabular}%
    }
  \label{tab:results_cmu}%
\end{table}%

%% file: 0-results-cmu-input.tex
\newcommand{\NetvladCMUUrbanRone}{12.0}
\newcommand{\NetvladCMUUrbanRfive}{30.5}
\newcommand{\NetvladCMUUrbanRten}{91.4}

\newcommand{\DensevladCMUUrbanRone}{15.0}
\newcommand{\DensevladCMUUrbanRfive}{37.2}
\newcommand{\DensevladCMUUrbanRten}{88.8}

\newcommand{\ApgemCMUUrbanRone}{\ 8.1}
\newcommand{\ApgemCMUUrbanRfive}{21.5}
\newcommand{\ApgemCMUUrbanRten}{81.3}

\newcommand{\SuperglueCMUUrbanRone}{17.1}
\newcommand{\SuperglueCMUUrbanRfive}{43.6}
\newcommand{\SuperglueCMUUrbanRten}{96.9}

\newcommand{\OursThreeCMUUrbanRone}{\textbf{19.2}}
\newcommand{\OursThreeCMUUrbanRfive}{\textbf{48.0}}
\newcommand{\OursThreeCMUUrbanRten}{\textbf{97.2}}

\newcommand{\NetvladCMUSuburbanRone}{3.9}
\newcommand{\NetvladCMUSuburbanRfive}{14.2}
\newcommand{\NetvladCMUSuburbanRten}{79.8}

\newcommand{\DensevladCMUSuburbanRone}{5.5}
\newcommand{\DensevladCMUSuburbanRfive}{20.2}
\newcommand{\DensevladCMUSuburbanRten}{84.3}

\newcommand{\ApgemCMUSuburbanRone}{2.4}
\newcommand{\ApgemCMUSuburbanRfive}{\,9.1\,}
\newcommand{\ApgemCMUSuburbanRten}{63.6}

\newcommand{\SuperglueCMUSuburbanRone}{5.6}
\newcommand{\SuperglueCMUSuburbanRfive}{21.6}
\newcommand{\SuperglueCMUSuburbanRten}{96.7}

\newcommand{\OursThreeCMUSuburbanRone}{\textbf{8.2}}
\newcommand{\OursThreeCMUSuburbanRfive}{\textbf{28.8}}
\newcommand{\OursThreeCMUSuburbanRten}{\textbf{97.0}}

\newcommand{\NetvladCMUParkRone}{2.7}
\newcommand{\NetvladCMUParkRfive}{11.2}
\newcommand{\NetvladCMUParkRten}{64.6}

\newcommand{\DensevladCMUParkRone}{5.5}
\newcommand{\DensevladCMUParkRfive}{21.3}
\newcommand{\DensevladCMUParkRten}{74.6}

\newcommand{\ApgemCMUParkRone}{1.7}
\newcommand{\ApgemCMUParkRfive}{6.6}
\newcommand{\ApgemCMUParkRten}{45.3}

\newcommand{\SuperglueCMUParkRone}{7.5}
\newcommand{\SuperglueCMUParkRfive}{30.5}
\newcommand{\SuperglueCMUParkRten}{\textbf{96.5}}

\newcommand{\OursThreeCMUParkRone}{\textbf{9.5}}
\newcommand{\OursThreeCMUParkRfive}{\textbf{34.9}}
\newcommand{\OursThreeCMUParkRten}{94.3}

%% file: 1-expanded-dataset-descriptions.tex
\section{Detailed Dataset Description}
\label{sec:datasets}
In this Section, we further detail the datasets that were introduced in Section 4.2 of the main paper. To recap, we evaluate Patch-NetVLAD on six of the key benchmark datasets: Nordland~\cite{sunderhauf2013we}, Pittsburgh~\cite{Torii2015repetitive}, Tokyo24/7~\cite{Torii2018}, Mapillary Streets~\cite{Warburg_2020_CVPR}, Oxford Seasons v2~\cite{RobotCar,toft2020long} and Extended CMU Seasons~\cite{CMUSeasons,toft2020long}. Collectively the datasets encompass a wide and challenging range of viewpoint change, appearance change and acqusition methods. 

\textbf{Nordland:}
The Nordland dataset~\cite{sunderhauf2013we} is recorded from a train traveling 728km through Norway in four different seasons. The four recordings are aligned frame-by-frame using available GPS information. We use the summer and winter traverses as the reference and query sets respectively, as these traverses have the highest appearance dissimilarity and are typically considered in the literature~\cite{camara2020visual,NP2013,HauslerHMPF,hausler2019multi}. As in~\cite{sunderhauf2013we}, we use the entire 728km traverse and subsample the video at 1 fps. Like previous works using this dataset~\cite{camara2020visual,SN2015,HauslerHMPF,hausler2019multi}, we remove all tunnels and times when the train is stopped, which resulted in 27,592 images for both reference and query sets.

\textbf{RobotCar Seasons v2:}
The RobotCar dataset~\cite{RobotCar} is a collection of traverses through Oxford, recorded with an autonomous car across multiple times of day and seasons. RobotCar Seasons v2~\cite{toft2020long} is a standardized benchmark subset of the RobotCar dataset, where the reference images are recorded in overcast conditions, and the query images are captured at a variety of times and conditions: dawn, dusk, sun, rain, overcast summer, overcast winter, snow, night and night-rain. Similarly to Nordland, RobotCar Seasons v2 mainly captures appearance changes, while viewpoint changes are relatively minor. 

An important detail of the structure of the RobotCar Seasons v2 dataset is that the reference images of the original RobotCar Seasons v2 dataset is split into 49 disjoint submaps which comprise the full traverse. Query images are captured from 17 of these submaps for all conditions and furthermore, for each query image the identity of the corresponding submap is provided. For our evaluation, we merge \emph{all} 49 submaps for the reference traverse and localize each query image against the whole merged reference traverse \emph{without using the given submap identity} provided with the dataset. This presents a substantially more challenging image retrieval task which showcases the difference between our proposed method and alternative approaches.

\textbf{Extended CMU Seasons:}
CMU Seasons~\cite{CMUSeasons} is similar to the RobotCar dataset: a car was driven around an 8.8km long route in Pittsburgh covering urban, residential, and park-like settings. Extended CMU Seasons~\cite{toft2020long} is a subset of the original CMU Seasons dataset that has been standardized for benchmark purposes. Extended CMU Seasons covers a single reference set and multiple query traverses under varying seasonal conditions spanning a one-year time frame. Unlike the RobotCar Seasons, CMU does not contain images that were captured at nighttime.

\textbf{Pittsburgh:}
The Pittsburgh dataset~\cite{Torii2015repetitive} contains 250k images collected via Google Street View. The reference and query images are captured at different times of the day and several years apart. For each place, 24 perspective images (two pitch and twelve yaw directions) are generated, which leads to high variations in both viewpoints and appearance. We use the Pitts 30k subset as described in~\cite{AR2018}, which contains 10k reference images and 6816 query images.

\textbf{Tokyo 24/7:}
The Tokyo 24/7 dataset~\cite{Torii2018} contains images at 125 distinct locations captured with smartphones at three different viewing directions and at three different times of the day. Contrary to the Nordland and Pittsburgh datasets, Tokyo 24/7 includes nighttime images.

\textbf{Mapillary Street Level Sequences (MSLS):}
The MSLS dataset~\cite{Warburg_2020_CVPR} has recently been introduced with the aim of facilitating lifelong place recognition research. It contains over 1.6 million images recorded in 30 major cities across the globe in urban and suburban areas over a period of 7 years. Compared to the other datasets, it includes variations in \emph{all} of the following: geographical diversity, season, time of day, viewpoint, and weather. Similarly to RobotCar Seasons v2 and Extended CMU Seasons, it contains a public validation set and a withheld test set. While the dataset is suited for sequence-based methods, we only evaluate the image-to-image task.